\documentclass[journal]{IEEEtran}
\makeatletter
\renewcommand{\@IEEEsectpunct}{\ \,}
\makeatother
\usepackage{cite}
\usepackage[dvips]{graphicx}
\usepackage{amsmath}
\usepackage{amsfonts}
\usepackage{amsopn}
\usepackage{amsthm}
\usepackage[linesnumbered,ruled]{algorithm2e}
\usepackage{array}
\usepackage{booktabs}
\usepackage[table]{xcolor}
\definecolor{Gray}{gray}{0.8}
\usepackage[caption=false,font=normalsize,labelfont=sf,textfont=sf]{subfig}
\usepackage{fixltx2e}
\usepackage{url}
\usepackage{microtype}
\usepackage{multirow}
\newtheorem{thm}{Theorem}

\usepackage{xcolor}
\definecolor{goldenpoppy}{rgb}{1, 0, 0}
\newcommand{\HL}[1] {#1}

\begin{document}
\title{Evolving Unsupervised Deep Neural Networks for Learning Meaningful Representations}
\author{Yanan~Sun,~\IEEEmembership{Member},~\IEEEmembership{IEEE},
        Gary~G.~Yen,~\IEEEmembership{Fellow},~\IEEEmembership{IEEE},
        and~Zhang~Yi,~\IEEEmembership{Fellow},~\IEEEmembership{IEEE}
        \thanks{This work is supported in part by the China Scholarship Council under Grant 201506240048;  in part by the Miaozi Project in Science and Technology Innovation Program of Sichuan Province under Grant 16-YCG061, China; in part by the National Natural Science Foundation of China for Distinguished Young Scholar under Grant 61622504; and in part by the National
Natural Science Foundation of China under Grant 61432012 and Grant U1435213.}
\thanks{Yanan Sun is with the College of Computer Science, Sichuan University, Chengdu 610065 CHINA and with the School of Engineering and Computer Science, Victoria University of Wellington, Wellington 6140 NEW ZEALAND. E-mail:yanan.sun@ecs.vuw.ac.nz.}
\thanks{Gary G. Yen is with the School of Electrical and Computer Engineering, Oklahoma State University, Stillwater, OK 74075 USA. E-mail:gyen@okstate.edu.~(\emph{Corresponding author})}
\thanks{Zhang Yi is with the College of Computer Science, Sichuan University, Chengdu 610065 CHINA. E-mail:zhangyi@scu.edu.cn.}
}

\maketitle

\begin{abstract}
Deep Learning (DL) aims at learning the \emph{meaningful representations}. A meaningful representation gives rise to significant performance improvement of associated Machine Learning (ML) tasks by replacing the raw data as the input. However, optimal architecture design and model parameter estimation in DL algorithms are widely considered to be intractable. Evolutionary algorithms are much preferable for complex and non-convex problems due to its inherent characteristics of gradient-free and insensitivity to \HL{the} local \HL{optimal}. In this paper, we propose a computationally economical algorithm for evolving \emph{unsupervised deep neural networks} to efficiently learn \emph{meaningful representations}, which is very suitable in the current Big Data era where sufficient labeled data for training is often expensive to acquire. In the proposed algorithm, finding an appropriate architecture and the initialized parameter values for \HL{an} ML task at hand is modeled by one computational efficient gene encoding approach, which is employed to effectively model the task with a large number of parameters. In addition, a local search strategy is incorporated to facilitate the exploitation search for further improving the performance. Furthermore, a small proportion labeled data is utilized during evolution search to guarantee the \HL{learned} representations to be meaningful. The performance of the proposed algorithm has been thoroughly investigated over classification tasks. Specifically, error classification rate on MNIST with $1.15\%$ is reached by the proposed algorithm consistently, which is \HL{considered} a very promising result against state-of-the-art unsupervised DL algorithms.
\end{abstract}

%
\IEEEpeerreviewmaketitle

\section{Introduction}
\label{section_1}
\IEEEPARstart{D}{eep} Learning (DL) algorithm, which is materialized by Deep Neural Networks (DNNs) for learning meaningful representations~\cite{bengio2013representation}, is a very hot research area during recent years~\cite{krizhevsky2012imagenet,farabet2013learning,tompson2014joint}.
Meaningful representation refers to the outcome of the raw input data that goes through multiple nonlinear transformations in the DNNs, and the outcome could remarkably enhance the performance of the subsequent machine learning tasks. The hyper-parameter settings and parameter values in DNNs are substantially interrelated to the performance of DL algorithms. Specifically, hyper-parameters \HL{(such as the size of weights, types of nonlinear activation functions, \HL{a} priori term types, and coefficient values)} refer to the parameters that are needed to be assigned prior to training the models, and parameter values refer to the element values of the weights and are determined during the training phase. Due to the deficiencies of the current optimization techniques for searching for optimal hyper-parameter settings and parameter values, the power of DL algorithms cannot be shown fully. To this end, an effective and efficient approach concerning the hyper-parameter settings and parameter values has been proposed in this paper.

\begin{figure}[htp]
  \centering
  \includegraphics[width=0.7\columnwidth]{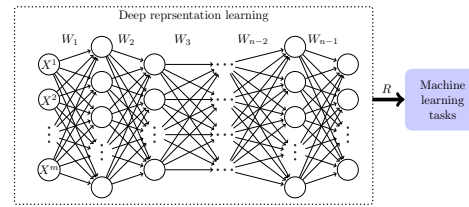}
  \caption{An example to illustrate a general \HL{flowchart} of deep representation learning and its relationship to machine learning tasks.}\label{fig_dl_general}
\end{figure}

\textbf{Meaningful Representations}~~Typically, arbitrary DNNs can generate/learn Deep Representations (DRs). However, DRs are not necessarily meaningful, i.e., it is not true that all DRs contributed to the promising performance when they replace the raw data to be fed to machine learning algorithms (e.g., classification). In fact, DRs are the outcomes which have gone through nonlinear transformations from input data more than once~\cite{delalleau2011shallow}, and are inspired by the mammalian hierarchical visual pathway~\cite{hubel1962receptive}. Mathematically, the representations of the input data $X\in \mathbb{R}^m$ are formulated by~(\ref{equ_deep_representation})
\begin{equation}
  \label{equ_deep_representation}
  \left\{
  \begin{array}{rl}
    R_1= & f_1(W_1X) \\
    R_2 =& f_2(W_2R_1) \\
     & \cdots \\
    R_n =& f_n(W_nR_{n-1}) \\
    R =& R_n
  \end{array}
  \right.
\end{equation}
where $f_1,\cdots,f_n$ denote a set of element-wise nonlinear activation functions, $W_1,\cdots, W_n$ refer to a series of connection weights and $R_1, R_2, \cdots,R_n$ are the \HL{learned} representations (output) at the depth/layer $1, 2, \cdots, $ and $n$, among which $\textbf{R}=\{R_i|2\le i\le n\}$ refers to the DRs. In addition, Fig.~\ref{fig_dl_general} shows the \HL{flowchart} of deep representation learning and its role in machine learning tasks in a general case.

Obviously, multiple different DRs can be \HL{learned} by varying $n$ in~(\ref{equ_deep_representation}), while we only pay attention to the ones that give the highest performance of the associated machine learning tasks. Based on literature reviews~\cite{bengio2009learning,sunlearning,sun2015explicit}, these DRs are often called \emph{meaningful representations}. Assuming $R_j$ are the meaningful representations, it is obvious that the hyper-parameter settings (e.g., the number of layers, $j$, and the chosen activation function types of $f_1,\cdots, f_j$) and parameter values (e.g., the values of each element in $\{W_1,\cdots, W_j\}$) would highly reflect the \HL{learned} $R_j$ to be meaningful or not. To this end, the Back-Propagation algorithm (BP)~\cite{rumelhart1988learning} which relies on the gradient information is the widely employed algorithm in training parameter values. However, its performance is highly affected by the initialized setting due to its local search characteristics that could be easily trapped into local minima~\cite{sutton1986two}. Although multiple implementations based on BP, such as Stochastic Gradient Descent (SGD), AdaGrad~\cite{duchi2011adaptive}, RMSProp~\cite{tieleman2012rmsprop}, and AdaDelta~\cite{zeiler2012adadelta}, have been presented to expectedly reduce the adverse impact of easily trapping into local minima, extra hyper-parameters (such as the initialization values of momentums and the balance factors) are introduced and also needed to be carefully tuned in advance. Furthermore, multiple algorithms~\cite{bergstra2011algorithms, bergstra2013making} have been proposed for optimizing the hyper-parameters, but they often require domain knowledge and are problem-dependent. To this end, the grid search method keeping its dominant position in selecting reasonable hyper-parameters was proposed~\cite{lerman1980fitting}. However, the grid search method is an exhaustive approach, and would frequently miss the best hyper-parameter combinations when the hyper-parameters are continuous numbers.

  \textbf{Deep Neural Networks}~~According to \HL{literature}~\cite{bengio2007scaling,lecun2015deep}, DL algorithms mainly include Convolutional NNs (CNNs), Deep Belief Networks (DBNs), and stacked Auto-Encoders (AEs). Specifically, CNNs are supervised algorithms for DL, and their numerous variants have been developed for various real-world applications~\cite{lecun1989backpropagation,lecun1998gradient,szegedy2015going,zeiler2014visualizing,simonyan2014very,he2015deep}. Although these CNN algorithms have shown promising performance in some tasks, sufficient labeled training data, which is a must for successfully training them, are not easy to acquire. For example in the ImageNet benchmark~\cite{deng2009imagenet}, there are $10^9$ pictures that can be easily downloaded from the Google and Yahoo websites. It was reported that $48,940$ workers from $167$ countries are employed to label these photos. Therefore, the unsupervised NN approaches whose training processes rely solely on unlabeled data become preferable in this situation. DBNs~\cite{hinton2006reducing} and stacked AEs~\cite{bourlard1988auto,hinton1994autoencoders} are the mainly unsupervised DL algorithms~\cite{bengio2007scaling,lecun2015deep} for learning meaningful representations. Because of the unknown in training data targets during their training phase, \HL{learned} representations from them are not necessarily to be \emph{meaningful}. Therefore, \textit{a priori} knowledge is needed to be incorporated into their training phase. For example, DBNs and stacked AEs trained with the sparsity constraint \HL{a} priori with benefits of sparse coding~\cite{olshausen1997sparse} have been proposed in~\cite{NIPS2007_3313} and~\cite{bengio2007greedy}. Furthermore, denoising AEs~\cite{vincent2008extracting} have been proposed by artificially adding noise priori to input data for improving the ability \HL{to learn} meaningful representations. In addition, Rifar \textit{et al.}~\cite{rifai2011contractive} have presented contractive AEs by introducing the term, which is the derivation of representations with respect to input data, for reducing the sensitivity \HL{a} priori of representations.

\textbf{Evolutionary Algorithms for NNs}~~Evolutionary algorithms (EAs) are one class of population-based meta-heuristic optimization paradigms, and are motivated by the metaphors of biological evolution. During the period of evolution, individuals interact with each other and the beneficial traits are passed down to facilitate population adapting to the environment. Due to the nature of \emph{gradient-free} and \emph{insensitivity to local optima}, EAs are preferred in various problem domains~\cite{yao1999evolving}. Therefore, they have been extensively employed in optimizing NNs, which refers to the discipline of neuroevolution, such as for the connection weight optimization~\cite{whitley1990genetic, whitley1989genitor, montana1989training}, the architecture setting~\cite{fahlman1989cascade,frean1990upstart,sietsma1991creating} (more examples can be found in~\cite{yao1999evolving}). Generally, these algorithms employ direct or indirect methods to encode the optimized problems for the evolution. To be specific, each parameter in the connection weights is encoded by the binary numbers~\cite{whitley1990genetic} or a real number~\cite{zi2007improvement} in the direct methods, which are effective for the \HL{small-scale} problems. However, when they are used to encode the problems with a large number of parameters in connection weights, such as for processing the high-dimensional data, these methods become impractical due to the excessive length of the genotype explicitly representing each parameter no matter if coded in binary or real. To this purpose, Stanley and Miikkulainen have proposed the indirect-based Neural Evolution Augmenting Topologies (NEAT) method~\cite{stanley2002evolving} for encoding connection weights and architectures with varying lengths of chromosomes. Because NEAT employs one unit to denote combinational information of one connection in the evolved NN, it still cannot effectively solve deep NNs where a large \HL{number} of parameters exist. To this end, an improved version of NEAT (i.e., HyperNEAT) was proposed in~\cite{stanley2007compositional} in which connection weights were evolved by composing different points in a fixed coordinate system with a series of predefined nonlinear functions. Although the indirect methods can reduce the length of the genotype representation, they limit the generalization of the neural networks and the feasible architecture space~\cite{yao1999evolving}. \HL{In 2015, Gong \textit{et al.}~\cite{gong2015multiobjective} proposed a bi-objective evolutionary algorithm by using Differential Evolution~\cite{Storn1997Differential} to concurrently consider the reconstruction error and sparsity of the AE, and chose the optimal sparsity from the knee area of the Pareto front.} Recently, Liu \textit{et al.}~\cite{liu2017structure} presented a neural network connection pruning method by a multi-objective evolutionary algorithm to simultaneously consider the representation ability and the sparse measurement. Google~\cite{real2017large} proposed their work on evolving CNNs for image classifications with a direct manner over $250$ high performance \HL{servers} for more than $400$ hours. In this regard, the evolutionary approaches would \HL{surely be capable of} evolving deep NNs, although \HL{the} computational resource is not necessarily available to all interested researchers.

\textbf{Contributions}~~Based on the above investigations upon prospects of unsupervised deep NNs for learning meaningful representations and the EAs in evolving deep NNs, an effective and efficient approach named Evolving Unsupervised Deep Neural Networks (EUDNN) for learning meaningful representation \HL{through evolving unsupervised deep NNs, exactly evolving the building blocks of unsupervised deep NNs, }has been proposed in this paper. In summary, the contributions of this paper are documented as follows:
\HL{\begin{enumerate}
  \item A computationally efficient gene encoding scheme of evolutionary approaches has been suggested, which is capable of evolving deep neural networks with a large number of parameters for addressing high-dimensional data with limited computational resources. With this design, the proposed algorithm can be smoothly implemented in academic environments with limited computational resources.
  \item A fitness evaluation strategy has been employed to drive the unsupervised models towards usefulness in advance, which can drive the \HL{learned} representations to be meaningful without any carefully designed \HL{a} priori knowledge.
  \item Deep neural networks with a large number of parameters involve a large-scale global optimization problem. As a result, \HL{the} sole evolutionary scheme cannot generate the best results. To this end, the utilization of a local search strategy is proposed to be incorporated into the proposed algorithm to guarantee the \HL{desired} performance.
\end{enumerate}}

\textbf{Organization}~~The remaining of this paper is organized as follows. First, related works and motivations of the proposed EUDNN are illustrated in Section~\ref{section_2}. Next, the details and discussions of the proposed algorithm are presented in Section~\ref{section_3}. To evaluate the performance, a series of experiments are performed by the proposed algorithm against selected peer competitors and the results measured by the chosen performance metric are analyzed in Section~\ref{section_4}. Finally, conclusions and future work are drawn in Section~\ref{section_5}.

\section{Related works and Motivations}
\label{section_2}
We will detail the unsupervised DL models that motive our work in this paper, highlight their deficiencies in learning \emph{meaningful} representations, and rationalize our motivations in Subsection~\ref{section_2_1}. With this same detailed manner, the evolutionary algorithms which demonstrate the potential for evolving deep NNs will be documented in Subsection~\ref{section_2_2}.

\subsection{Unsupervised Deep Learning Models}
\label{section_2_1}
In this subsection, the unsupervised DL models are reviewed first (Subsection~\ref{section_2_1_1}). Then, their building blocks are introduced (Subsection~\ref{section_2_1_2}). Next, the mechanisms guaranteeing the \HL{learned} representations to be meaningful are formulated and commented (Subsection~\ref{section_2_1_3}). Finally, the motivations of the proposed algorithm in reducing the adverse impact of their deficiencies are elaborated (Subsection~\ref{section_2_1_4}).

\subsubsection{}\label{section_2_1_1}
\begin{figure}[htp]
\begin{center}
\subfloat[pre-training]{\includegraphics[width=0.8\columnwidth]{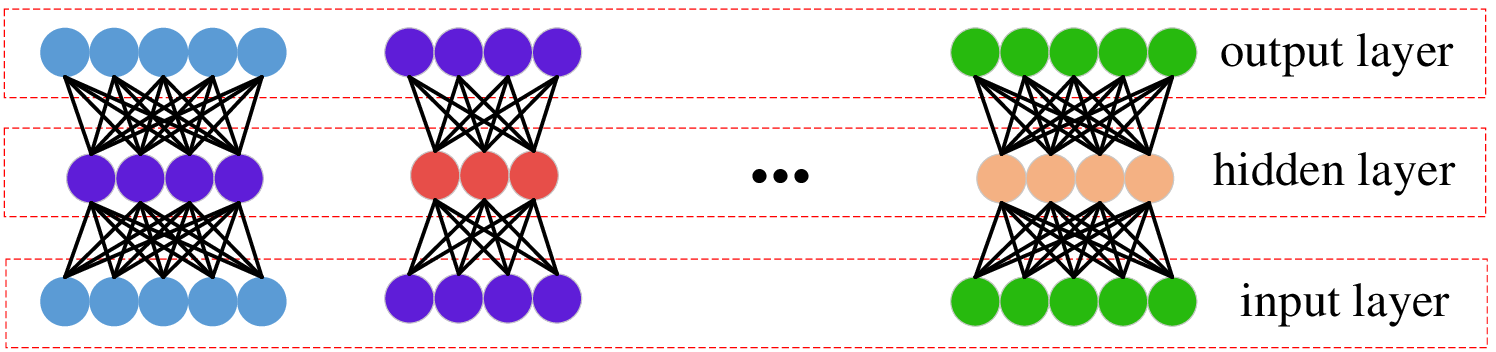}%
\label{fig_dl_training_1}}
\hfil
\subfloat[fine-tuning]{\includegraphics[width=0.8\columnwidth]{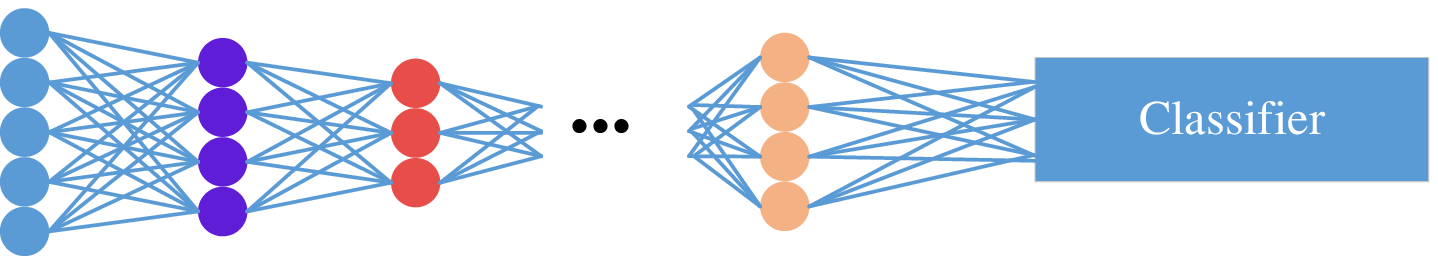}%
\label{fig_dl_training_2}}

\caption{\HL{The training} process of unsupervised deep neural networks.}
\label{fig_dl_training}
\end{center}
\end{figure}

Unsupervised DL models cover DBNs~\cite{hinton2006reducing} and variants of stacked AEs (i.e., stacked sparse AEs (SAEs)~\cite{NIPS2007_3313,bengio2007greedy}, stacked denoising AEs (DAEs)~\cite{vincent2008extracting}, and stacked contract AEs (CAEs)~\cite{rifai2011contractive}). Moreover, the building block of DBNs is a Restricted Boltzmann machine (RBM)~\cite{smolensky1986information}, and that of stacked AEs is an AE. Furthermore, the parameter values in DBNs and stacked AEs are optimized by the greedy layer-wise training method, which is composed of two phases~\cite{hinton2006fast}: pre-training and fine-tuning. Conveniently, Fig.~\ref{fig_dl_training_1} depicts the pre-training phase, where a set of three-layer (the input layer, the hidden layer, and the output layer) NNs with varying numbers of units are individually trained by minimizing reconstruction errors. In the fine-tuning phase which is illustrated by Fig.~\ref{fig_dl_training_2}, these hidden layers are first sequentially stacked together with the parameter values trained in the pre-training phase, then a classification layer (i.e., the classifier) is added to the \HL{tail} to perform the fine-tuning by optimizing the corresponding loss function determined by the particular task at hand.

\subsubsection{}\label{section_2_1_2}
\begin{figure}[htp]
  \centering
  \includegraphics[width=0.6\columnwidth]{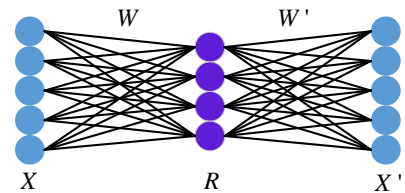}\\
  \caption{An example of unsupervised deep neural network unit model.}\label{fig_dl_unit}
\end{figure}

Unsupervised DL algorithms are considerably preferred mainly due to their requirements upon fewer labeled data especially in the current Big Data era\footnote{Even data is abundant in the Big Data era, most raw data collected is unlabeled for a classification task, e.g., the ImageNet classification benchmark that has been discussed in Section~\ref{section_1}.}. However, a major issue of training these models is how to guarantee the \HL{learned} representations to be meaningful. Specifically in the pre-training phase for training one NN unit (see Fig.~\ref{fig_dl_unit} as an example), let $X\in R^n$ denote the input data, \HL{$W\in R^{n\times k}$} denote the connection weight matrix from the input layer to the hidden layer, while $ W'\in R^{k\times n}$ denote the connection weight matrix from the hidden layer to the output layer. The NN unit is trying to minimize the reconstruction error $L$ between the input data $X$ and the output data $X'$ by~(\ref{equ_dl_unit})\footnote{Bias terms, which are another kind of connection weights widely existing in NNs, are incorporated into $W$ and $W'$ here for simplicity.}

\begin{equation}
  \label{equ_dl_unit}
  \left\{
  \begin{array}{lll}
    R &= &f(WX) \\
    X' &= &f(W'R) \\
    L &= &l(X,X')
  \end{array}
  \right.
\end{equation}
In~(\ref{equ_dl_unit}), $R$ denotes the \HL{learned} representations (i.e., the output of the hidden layer), $f$ denotes the activation function, and $l$ denotes the function to measure the differences between $X$ and $X'$.
\HL{
\subsubsection{}\label{section_2_1_3}
It is obvious that the \HL{learned} representations $R$ are not necessarily meaningful only by minimizing $L$ due to no information of the associated classification task existing in this phase and arbitrary $R$ will lead to a minimal $L$, while $R$ is meaningful only when they could improve the performance of the associated classification task. To this end, \HL{literature} have presented unsupervised DL algorithms with different \HL{a} priori knowledge~\cite{olshausen1997sparse,bengio2007greedy,vincent2008extracting,rifai2011contractive} which is denoted as $\Theta$, and then the reconstruction error is transformed to $L=l(X,X')+\lambda \Theta$ where $\lambda$ denotes a balance factor to determine the weight of the associated \HL{a} priori term. Although \HL{a} prior knowledge would help the \HL{learned} representations to be meaningful, major issues remain:
 \begin{itemize}
   \item The prior knowledge is designed with different assumptions, which do not necessarily satisfy the current situations.

   \item The prior knowledge is presented specifically for general tasks, while it is hopeful that the performance would be improved on particular tasks.

   \item It is difficult to choose the most suitable a priori term for the current task.

   \item The balance factor $\lambda$ is a hyper-parameter whose value is not easily to be assigned~\cite{rifai2011contractive}.
 \end{itemize}
}

\HL{
\subsubsection{}
\label{section_2_1_4}
Considering this problem, the method that has been developed in our previous work~\cite{sun2015explicit} is employed in this proposed algorithm. To be specific, a small proportion of labeled data is employed during the fitness evaluation of EAs, and the \HL{learned} representations are directly quantified based on the classification task that is employed in the fine-tuning phase. With the environmental selection in EAs, individuals that have the positive effect on the classification task survive into the current generation and are \HL{expected} to generate offspring with better performance in the next generation, which \HL{ultimately} leads to the \HL{learned} representations to be meaningful. Because the employed labeled data can be injected from the fine-tuning phase, and the classification task is the same as that in the fine-tuning phase, this strategy for learning meaningful representations would not introduce extra cost.
}

\subsection{Evolutionary Algorithms for Evolving Neural Networks}
\label{section_2_2}
Although multiple related \HL{literature} for evolving NNs have been mentioned in Section~\ref{section_1}, only the works in~\cite{stanley2002evolving,stanley2007compositional} (i.e., the NEAT and the HyperNEAT) will be concerned here because our proposed algorithm aims at evolving \emph{deep} NNs\footnote{The works in~\cite{
whitley1990genetic, whitley1989genitor, montana1989training,fahlman1989cascade,frean1990upstart,sietsma1991creating,yao1999evolving} were proposed two decades ago and cannot be applied for deep NNs, the work in~\cite{liu2017structure} concerned only the weight pruning, and the work in~\cite{real2017large} employed a direct way for evolving and did not have a general meaning.}. In the following, the details of NEAT\HL{,} as well as HyperNEAT and their deficiencies in evolving deep NNs are documented in Subsections~\ref{section_2_2_1} and~\ref{section_2_2_2}, respectively. Combined with the challenge of EAs in evolving deep NNs, i.e., the upper bound encoding problem, the motivations of the proposed EUDNN are presented in Subsection~\ref{section_2_2_3}. In addition, another challenge, i.e., EAs cannot fully solve the optimization problems with a large number of parameters, and the corresponding motivations are given in Subsection~\ref{section_2_2_4}.

\subsubsection{}\label{section_2_2_1}
The NEAT~\cite{stanley2002evolving} has been proposed with an indirect method for adaptively increasing the complexity of the evolved NNs. Specifically, two types of genes, i.e., the node genes and the connection genes, exist in the NEAT. The node genes, which are used to represent all the units of the evolved NN, are encoded with the type of the unit (i.e., the input unit, the hidden unit, or the output unit) and one identification number. The connection genes that are employed to denote the connection information between the node genes, and one node gene is encoded with five elements (the numbers of the input and output units, the value of the connection, one bit indicating whether the connection is activated or not, and one innovation number which records the index of the connection gene with an increased manner). During the evolution process, the individuals are first initialized only with the input and output units of the network, and the random connections between these units. Then, individuals are recombined and mutated. To be specific, there are two types of mutations including the connection mutations and the node mutations. When the connection mutations occur, one connection gene will be added to the list of the connection genes to denote that a pair of node genes is connected. While for the node mutations, one hidden node is generated, then the corresponding connection gene is created to split one existed connection into two parts. Although the NEAT is flexible to evolve NNs, a deterministic number of the output is required, which is impractical in the DL. Furthermore, due to each connection and unit in NEAT are explicitly encoded, it is not suitable for evolving deep NNs that often have a large number connections and units. For remedying this deficiency regarding the incapacity of evolving deep NNs, the connective compositional pattern producing networks (CPNN)~\cite{stanley2006exploiting, stanley2007compositional} has been presented and led to the HyperNEAT.

\subsubsection{}\label{section_2_2_2}
The HyperNEAT has been proposed by combining the NEAT with the CPNN encoding scheme. Particularly, the CPNN employs one low-dimensional coordinate system to generate connections for the NEAT by a list of predefined nonlinear functions. To be specific, any point in the coordinate system is picked up, and then fed into a series of compositional functions from the list to complete the transformation from the genotype to the phenotype. Because any number of points can be selected from the low-dimensional coordinate system, numerous connections would be represented with a low computational cost. In this regard, the HyperNEAT has the most potential for evolving a deep NN, while the size of the output still needs to be set in advance, which \HL{faces} the same problem to NEAT in practice. Furthermore, all the values of the connections in the HyperNEAT are generated by the genetic operators during the evolution, which cannot guarantee the best performance in evolving a deep NN due to the nature of the large-scale global problem. In addition, the recurrent connections or the connections between the same layers are involved in this algorithm, which \HL{is} also not suitable for learning compact meaningful representations.
\subsubsection{}\label{section_2_2_3}
 As we have discussed in Section~\ref{section_1}, the performance of DL algorithms is highly affected by the hyper-parameter settings and the parameter values. In the pre-training phases, one of the key hyper-parameters is the size of hidden layers. One problem would be naturally raised when EA approaches are employed to search for the sizes, that is how we can ensure the upper bound of the hidden layer sizes given a fixed-length gene encoding strategy. Although the indirect encoding scheme can alleviate this situation somewhat, it limits the generalization of the evolved NNs and the feasible architecture space~\cite{whitley1990genetic}. On the other hand, if we employ a larger number as the upper bound, it is difficult to determine how large it is reasonable because too large a number would consume more computational resources, otherwise deteriorate the model performance. Excitingly, Yang \emph{et al.}~\cite{yang2005kpca} have mathematically pointed out that the meaningful representations of the input data lie at its original space. Supposed that the input data is with $n$ dimension, the size of \HL{the} associated hidden layer should be no more than $n$. Furthermore, we know that $n$ orthogonal $n$-dimensional basis vectors are sufficient to span a $n$-dimensional space \HL{based on Theorem~\ref{the_1}}. Consequently, we only need to compute one basis $r_1$ of $n$-dimensional space, and the other $(n-1)$ $n$-dimensional basis vectors can be explicitly computed by~(\ref{equ_null_space}) to find the null space\HL{\footnote{\HL{Theoretically, multiple solutions could be found in computing the bases of the null space. In practice, we only accept the orthonormal basis for the corresponding null space obtained from the singular value decomposition.}}}. To this end, we can efficiently model the problem with $n^2$ parameters by employing a genetic algorithm to explicitly encode about $n$ parameters, which is a computational efficient gene encoding approach.
 \HL{
 	\begin{thm}
 		\label{the_1}
 		A set of orthogonal vectors $b_i\in R^n$ ($i=1,\cdots, n$) is sufficient to span the space $S\in R^n$.
 		
 	\end{thm}
 }
\begin{equation}
\label{equ_null_space}
  \text{null space}(r_1) = \{x\in R^n| r_1x=0\}
\end{equation}

\subsubsection{}\label{section_2_2_4}
Here, we would point out another challenge to inspire our motivation for evolving deep NNs by employing GAs. In our proposed algorithm, the computationally efficient gene encoding strategy mentioned above is employed to model unsupervised deep NNs where a large number of parameters exist. Although the length of the encoded parameters has been reduced appreciably in this regard, the number of the parameters in the original problems remains constant no matter what encoding method is employed. In fact, the effects of one gene in the employed encoding strategy is equivalent to that of multiple parameters in the original problems. For example, for \HL{an} NN which has $100,000$ parameters, only $1,000$ genes are employed by the computationally efficient gene encoding strategy proposed herein. As a result, one gene represents $100$ parameters in average, and if one gene is changed with the crossover and mutation operators, it will involve the changes of $100$ parameters. Moreover, it is well known that performances of EAs are guaranteed by their exploration search (given by mutation operators) and exploitation search (given by crossover operators) which introduce the global search and local search abilities, respectively. Because a slight change of one gene in the proposed algorithm will lead to the changes of many parameters which affect the global behavior, it can be viewed as that EAs lack of the local search from the problem to be solved. In addition, the data which are processed by DL algorithms is \HL{common} with high dimension, which leads to a large number of decision variables in the encoded chromosomes of EAs, although our employed encoding strategy has saved much space compared to existing approaches. Extensive experiments have quantified that EAs are difficult to reach the best performance upon the problems with high input dimensions. To address this issue, we incorporate a local search strategy into the proposed algorithm for assuring the desirable performance.

In summary, the difficulties of deep unsupervised NNs for learning meaningful representations and EAs for evolving deep NNs have been clarified first, and then addressed by our motivations in this section. In the next section, the technical details will be implemented based on these motivations.
\section{Proposed Algorithm}
\label{section_3}
In this section, the details of the proposed EUDNN are presented. To be specific, the framework which is composed of two distinct stages is depicted at first (Subsection~\ref{section3_1}). Next the specifics of each stage are elaborated, respectively (Subsections~\ref{section3_2} and~\ref{section3_3}). Furthermore, the over-fitting problem preventing mechanism of EUDNN and the significant differences against its peer competitor are discussed (Subsection~\ref{section3_4}).

\begin{figure*}
  \centering
  \includegraphics[width=0.85\textwidth]{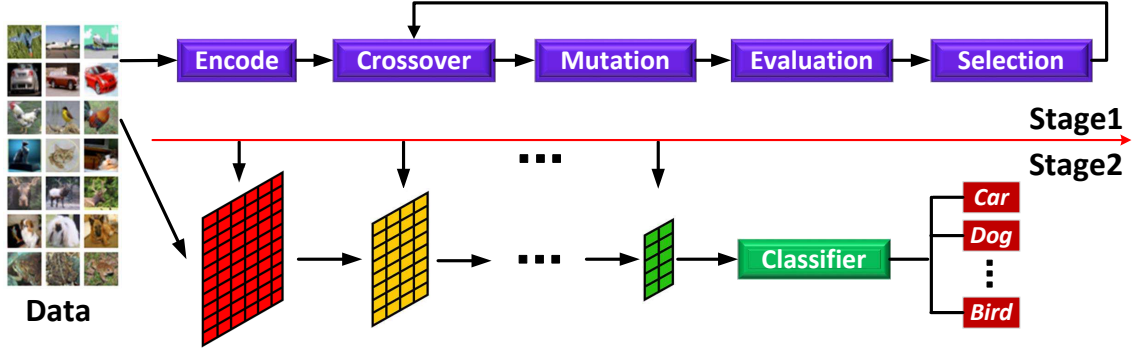}\\
  \caption{The \HL{flowchart} of the proposed algorithm that is composed of two distinct stages. Especially, the first stage is for finding optimal architectures as well as \HL{desirable} initializations of the connection weight parameter values. The second stage is to fine-tune them for a \HL{potentially} better performance.}\label{fig_flowchart}
\end{figure*}

\subsection{Framework of EUDNN}
\label{section3_1}
In this subsection, the framework of the proposed EUDNN is presented. For convenience of the development, it is assuming that the \HL{learned} representations are for a classification task in which the \emph{meaningful} representations can improve its performance in term of a higher Correct Classification Rate (CCR) (the CCR upon the training data is collected during the training/optimization phase, and that upon the test data during the test/experimental phase). Moreover, given a set of data $D$ in this classification task, a portion of $D$ which is denoted by $D_{train}=\{(x_1,y_1),\cdots,(x_k,y_k)\}$ is considered as the training data in which $x_i$ denotes the input data and $y_i$ is the corresponding label, while the remaining data is regarded as the test data $D_{test}$ for checking whether the \HL{learned} representations are meaningful. Furthermore, the \HL{flowchart} of the proposed EUDNN is illustrated in Fig.~\ref{fig_flowchart}, which clearly shows the two stages of the design: 1) finding the optimal architectures in deep NNs, the desirable initialization of connection weight, and the activation functions (pre-training), and 2) fine-tuning all of the parameter values in connection weights from the desirable initialization.

\begin{algorithm}
  \caption{Framework of the Proposed EUDNN}
  \label{alg_the_proposed_algorithm}
    \KwIn{Training data $D_{train}$; maximum number $p$ of layers; classifier $C(\cdot)$; test data $D_{test}$.}
    \KwOut{Predicted labels of $D_{test}$.}
    $i\leftarrow 0$;\\
    \While{$i<p$}
    {
    \label{alg_framework_stage1_begin}
        $i\leftarrow i+1$;\\
        $W_j, f_j(\cdot) \leftarrow$ Obtain the optimal connection weight and the corresponding activation function via evolving;\\
    }
    \label{alg_framework_stage1_end}
    Fine-tune all the connection weights $W_1,\cdots, W_p$;\\
    \label{alg_framework_stage2}
    $Y_{test}=C(f_p(W_p\times \cdots f_2(W_2\times f_1(W_1\times D_{test}))))$;\\
    \label{alg_framework_predict_label}
    \textbf{Return} $Y_{test}$.
    \label{alg_framework_return_label}
\end{algorithm}

To this end, one genetic approach with an efficient strategy introduced in Subsection~\ref{section_2_2} is employed to encode the potential architectures and the associated large numbers of parameters in connection weights by a set of individuals, and then the EA is utilized to evolve and select the individual who has the best performance based on the fitness measures. For warranting the \HL{learned} representations being meaningful, the method introduced in Subsection~\ref{section_2_1} is employed, i.e., a small part of data $D_f$ from $D_{train}$ is randomly selected, and the representations of $D_f$ are \HL{learned} based on the models encoded by the individuals, then \HL{they} are fed with the associated classification task to select the ones which give the higher CCR for evolution. Based on the investigations in Subsection~\ref{section_2_2}, a fine-tuning approach additionally, which introduces the exploitation local search, is utilized in the second stage to archive the best performance ever found, which complements with the exploration global search in the first stage. In summary, these two stages collectively ensure the \HL{learned} representations to be meaningful through unsupervised deep NNs.

In addition, the framework of the proposed EUDNN is presented in Algorithm~\ref{alg_the_proposed_algorithm}. Specifically, lines~\ref{alg_framework_stage1_begin}-\ref{alg_framework_stage1_end} describe the first stage, while line~\ref{alg_framework_stage2} defines the second stage. Finally, the predicted labels of the test data are calculated and returned in lines~\ref{alg_framework_predict_label} and~\ref{alg_framework_return_label}. Next, the details of these two stages are documented, respectively.

\subsection{Obtaining Optimal Connection Weights and Activation Functions via Evolving}
\label{section3_2}
The process of obtaining all the optimal connection weights and their corresponding activation functions contains a series of repeated subprocesses. In this subsection, we first in Algorithm~\ref{alg_obtain_transformation_matrix} propose how to obtain one optimal connection weight and its activation function. Then, the entire process is described.
\begin{algorithm}
  \caption{Obtain the Optimal Connection Weight and Activation Function}
  \label{alg_obtain_transformation_matrix}
    \KwIn{Input data; size of population $m$; probability of crossover $\rho$; probability of mutation $\mu$.}
    \KwOut{Optimal connection weight $W$; activation function $f(\cdot)$.}
    Initialize the population $P$ with the size $m$;\\
    \label{alg_obtain_transformation_matrix_line1}
    \While{stopping criteria are not satisfied}
    {
        \label{alg_otme_begin}
        Evaluate the fitness of individuals in $P$;\\
        \label{alg_obtain_transformation_matrix_line2}
        $Q\leftarrow$ Generate new offspring with the probability $\rho$ from two parents selected with binary tournament selection;\\
        \label{alg_obtain_transformation_matrix_line3}
        $Q\leftarrow$ Mutate all the individuals in $Q$ with the probability $\mu$;\\
        \label{alg_obtain_transformation_matrix_line4}
        $S\leftarrow$ Select the individual with the best fitness from $P\cup Q$;\\
        \label{alg_obtain_transformation_matrix_line5}
        $P\leftarrow$ $S~\cup$~Select $(m-1)$ individuals from $(P\cup Q)\setminus S$ \HL{with binary tournament
selection};
        \label{alg_obtain_transformation_matrix_line6}
    }
    \label{alg_otme_end}
    Evaluate the fitness of the individuals in $P$;\\
    \label{alg_obtain_transformation_matrix_line7}
    $ind_{best}\leftarrow$ Select the individual with the best fitness from $P$;\\
    \label{alg_obtain_transformation_matrix_line8}
    \textbf{Return} $W$ and $f(\cdot)$ represented by $ind_{best}$.
\end{algorithm}

To be specific in Algorithm~\ref{alg_obtain_transformation_matrix}, $m$ individuals that encode the information of potential optimal connection weights and their corresponding activation functions are initialized first (line~\ref{alg_obtain_transformation_matrix_line1}). Then, the evolution takes effect (lines~\ref{alg_otme_begin}-\ref{alg_otme_end}) until the stopping conditions, such as exceeding the maximum generations, are met. During each generation, the fitness of all the individuals are evaluated first (line~\ref{alg_obtain_transformation_matrix_line2}). Next, new offspring are generated with the probability $\rho$, and their parents are selected from $P$ with the binary tournament selection (line~\ref{alg_obtain_transformation_matrix_line3}). Then, all the offspring in $Q$ are mutated with the probability $\mu$ (line~\ref{alg_obtain_transformation_matrix_line4}). Furthermore, lines~\ref{alg_obtain_transformation_matrix_line5}-\ref{alg_obtain_transformation_matrix_line6} describe the environmental selection in which the best individual is preserved first for the elitism\HL{, then $m-1$ individuals are selected from the remaining solutions in $P\cup Q$ with binary tournament selection. Specifically, two individuals are randomly selected from $(P\cup Q)\setminus S$ first. Then the one with better CCR is chosen, and the other is put back. With the same process, this operation is repeated $m-1$ times}.

\begin{figure}
  \centering
  \includegraphics[width=\columnwidth]{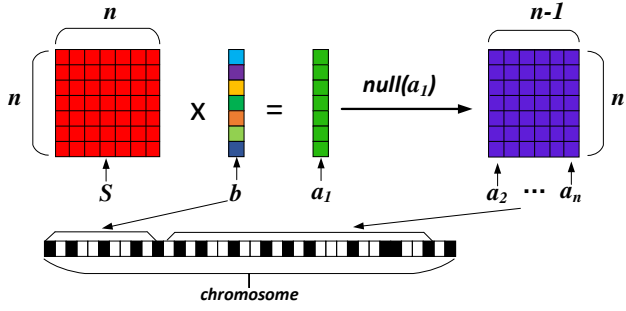}\\
  \caption{A \HL{flowchart} describes the process of encoding the potential connection weight and activation function. First, a set of basis vectors $S$ is given in the original space with $n$-dimension. Then, a set of coefficients $b$ is generated to represent the vector $a_1$ by linear combining the basis vectors. Then, the orthogonal complements $\{a_2,\cdots,a_n\}$ of $a_1$ are computed. Finally, all the information of computing $a_1$, indicating whether the basis from $\{a_2, \cdots, a_n\}$ is selected, and the activation functions are encoded into the chromosomes that are used to evolve to obtain the optimal connection weight and activation function.}\label{fig_encode_transformation_matrix}
\end{figure}

When the evolution terminates, the best solution is selected from the current population for transforming the optimal connection weight and the activation function (lines~\ref{alg_obtain_transformation_matrix_line7}-\ref{alg_obtain_transformation_matrix_line8}).
Next, the details of the employed gene encoding strategy will be discussed, although its fundamental principles have been documented in Subsection~\ref{section_2_2}. It has been pointed out in~\cite{yang2005kpca} that the potential connection weight for obtaining the meaningful representations likely lies in a subspace of the original space. As a consequence, the search for the optimal connection weight can be constrained in the space of input data. Specifically, it is assuming that the input data is $n$-dimensional. First, a set of basis $S=[s_1,\cdots,s_n]$ which can span a $n$-dimensional space is given, e.g., any $n$ linear independent $n$-dimensional vectors. Then the vector $a_1$ is linearly combined by the bases in $S$ with the coefficients $b=[b_1,\cdots,b_n]$ \HL{that are randomly specified}. Next, the orthogonal complements $\{a_2,\cdots,a_n\}$ of $a_1$ are computed by~(\ref{equ_null_space}). It is obvious that $\{a_1,a_2,\cdots,a_n\}$ are capable of spanning the space of input data. Finally, a part of these bases, which span a subspace of the original space, are selected for constructing the optimal connection weight by a binary encoded string indicating whether the corresponding basis is available. Furthermore, the corresponding activation function is also encoded into the chromosome. Specifically, a list of selected activation functions with different nonlinear capacities is given, then their indexes in this list are chosen to indicate which one is selected. Moreover, Fig.~\ref{fig_encode_transformation_matrix} is provided to intuitively illustrate our intention on efficiently encoding the connection weight and activation function. When the optimal connection weight $W_i$ and its corresponding activation function $f_i$ are found for the $i$-th layer with Algorithm~\ref{alg_obtain_transformation_matrix}, then that for the $(i+1)$-th layer can be optimized with the same algorithm by setting the input data as $f_i(W_i\times R_i)$ where $R_i$ denote the representations at the $i$-th layer.
\begin{figure*}[htp]
  \centering
  \includegraphics[width=\textwidth]{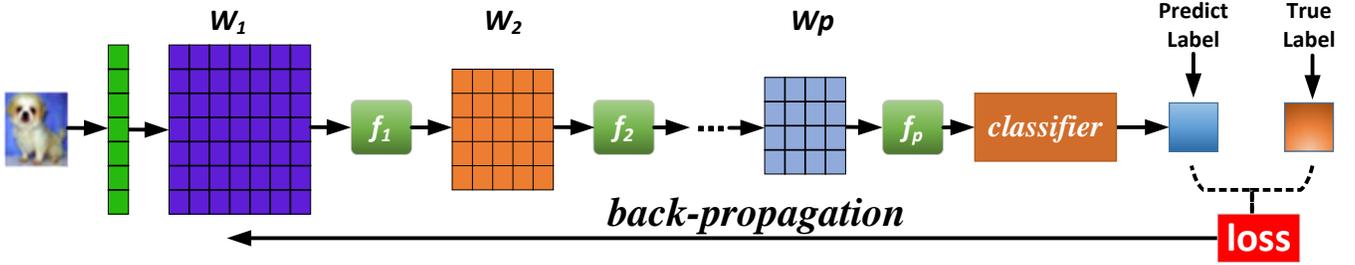}\\
  \caption{The \HL{flowchart} of the second stage in the proposed EUDNN. Especially, the predicted label is computed with the connection weights and activation functions for the input data. Then the loss of the classifier is formulated between the predicted label and the true label. Next, the error is back propagated and the parameter values of the connection weights are updated. }\label{fig_finetune}
\end{figure*}
In the employed gene encoding approach, each coefficient \HL{of} $b$ is represented with nine bits in which the \HL{leftmost} bit denotes the positive or negative of the coefficient. Then, one bit is used to indicate whether the basis $a_j$ ($j\in[2,\cdots,n]$) is selected for the connection weight. Finally, two bits are utilized to represent the activation function. In addition to the well-adopted sigmoid and hyperbolic tangent functions, rectifier function~\cite{glorot2011deep}, which is reported recently to have a superior performance in some applications, is also considered as one candidate. As a consequence, one chromosome needs $10n+1$ bits for the $n$-dimensional input data. If the real number encoding method is employed here, a multiple of eight memory space would be taken, which is the major reason that the proposed EUDNN employs the binary encoding method being \HL{a} contribution to the so claimed computational efficient gene encoding strategy.

Furthermore, the linear Support Vector Machine (SVM)~\cite{cortes1995support} is employed for evaluating the quality of individuals due to its promising computational efficiency and its linear nature for better discriminating power whether the \HL{learned} representations are meaningful or not. \HL{Next, we will give the details of the fitness evaluation by using SVM based on the design principle described in Subsection~\ref{section_2_1_4}.  For convenience of the development, let $D_{train}=\{X_{train}, Y_{train}\}$ denote the training set where $X_{train}$ are the data and $Y_{train}$ are the corresponding labels, and the selected individual for fitness evaluation is denoted by $ind_{i}$. Firstly, a small fraction of data denoted by $D_{eval}=\{X_{eval}, Y_{eval}\}$ is randomly selected from $D_{train}$. Secondly, the corresponding model is transformed from the encoded individual $ind_{i}$. Thirdly, the representations  (denoted by $F_{eval}$) of $X_{eval}$ are calculated based on the formulas in (\ref{equ_deep_representation}). Fourthly, $\{F_{eval}, Y_{eval}\}$ are fed to SVM and the CCR on $X_{eval}$ is estimated. Finally, the CCR is used as the fitness of $ind_i$.}

\subsection{Fine-tuning Connection Weights}
\label{section3_3}
\HL{To further improve the performance, an exploitation mechanism implemented by local search strategy is incorporated into the second stage to fine-tune parameter values in connection weights. In this stage, the architecture is fixed with the evolved activation functions and the initialization values of the connection weights, and then a local search method is used to tune the connection weights further. Fig.~\ref{fig_finetune} shows an example of this process. Specifically, when all the connection weights and activation functions have been optimized in the first stage, all the hidden layers are connected to a list based on their orders in the first stage by adding one input layer at the top of this list. Then, the connection weights in this list are initialized with the values confirmed in the first stage. Finally, a classifier is added to the tail of this list to perform the fine-tuning process. Note here that the BP algorithm is employed for the fine-tuning. Actually, any local search algorithm can be used in the second stage. The reasons \HL{for} employing BP are largely due to two aspects: 1) the gradient information in the loss function is always analytical and the BP that is based on the gradient is naturally employed in most designs; 2) multiple libraries of BP have been implemented for accelerating the computation with the Graphics Processing Units (GPUs) and the computational cost can be reduced remarkably, especially in the situations of processing high-dimensional data. Furthermore, when the rectifier activation function that is not differentiable at the point $0$ is selected, the value $0$ is assigned according to the convention of the community~\cite{maas2013rectifier}.}

\subsection{Discussions}
\label{section3_4}
In this subsection, we mainly discuss the over-fitting problem preventing mechanism utilized by the proposed EUDNN, and the significant differences of the proposed EUDNN against the Direct Evolutionary Feature Extraction algorithm (DEFE)~\cite{zhao2006direct} that employs a similar gene encoding strategy to EUDNN.

The over-fitting problem implies the poor generalization ability of models, i.e., the trained model reaches a better CCR upon training data at the cost of a worsen CCR upon test data. Because the goal in training a classification model is for obtaining a higher CCR upon test data, the over-fitting problem should be prevented by some mechanisms. Commonly, given a number of models which are all capable of solving a particular classification task, the model with a smaller Vapnik Chervonenkis (VC) dimension\footnote{Generally, the VC dimension can be viewed as an indicator measuring the complexity of multiple models which are capable of solving one particular task~\cite{Vapnik1997The}. The smaller the VC dimension, the more simplicity is the corresponding model, and a more simplicity model is with better generalization~\cite{Vapnik2010Statistical}. Commonly, a large number and magnitude of elements in the transformation matrixes are positive to the VC dimension.}~\cite{bengio2009learning} usually has a better generalization ability, which does not lead to an over-fitting problem. Because the number of parameters \HL{is} positive to the value of a VC dimension, and deep NN architectures are generally with \HL{the} numerous number of parameters, the over-fitting problem easily \HL{occurs} in these models.
\begin{figure}[htp]
  \centering
  \includegraphics[width=0.85\columnwidth]{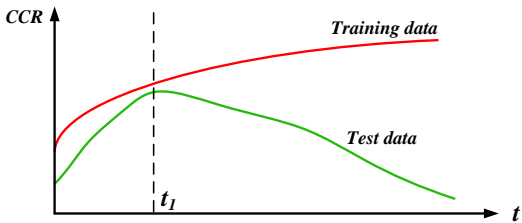}\\
  \caption{Correct classification rates of training data and test data as training process continues.}\label{fig_overfitting}
\end{figure}

More specifically, Fig.~\ref{fig_overfitting} illustrates a typical instance in CCR on training data (red curve) and CCR of checking on test data (green curve) as the training process continues. Especially, CCR on both data are continuously growing until the time $t_1$, and CCR on the training data continues to increase while CCR on the test data begins to drop when the training time is greater than $t_1$, which obviously indicates the presence of an over-fitting problem. As we have claimed that the best performance of the proposed EUDNN cannot be guaranteed during the training in the first stage, and the second stage is introduced to expectedly help the proposed EUDNN arrive at the best performance. To this end, it is concluded that the over-fitting problem will not occur in the first stage of the proposed EUDNN (because the first stage terminates prior to the time $t_1$, while the over-fitting problem might occur after the time $t_1$), but may occur in the second stage. Consequently, some rules need to be utilized to prevent this problem only in the second stage. Here, the ``early stop'' approach is utilized for this purpose, i.e., a group of data $D_{validate}$ is uniformly selected from $D_{train}$ as the validate data to replace the checking upon test data in Fig.~\ref{fig_overfitting}, when we first observe the CCR of validate data begins to decrease while the CCR of training is still increasing (i.e., the particular time $t_1$ is found), the fine-tuning in the second stage is terminated and the optimal model that gives the best performance is obtained. Next, the second concern, i.e., the differences between the proposed EUDNN and the DEFE, will be discussed.

It has been observed that 1) DEFE learns only linear representations and 2) shallow representations of input data. These two observations cause that DEFE cannot learn the meaningful representations~\cite{hinton2006reducing}. Next, the details of these conclusions are discussed. To be specific, the \HL{learned} representations $R$ of DEFE can be formulated as $R=WX$~\cite{zhao2006direct} where $W$ is the transformation matrix (i.e., the connection weight in deep NN models) and $X$ is the input data. It is evident that there is no nonlinear transformation upon $WX$. Consequently, only linear representations would be \HL{learned} by DEFE, while in the proposed EUDNN, a list of nonlinear activation functions with different nonlinear transformation abilities \HL{is} incorporated into the evolution for performing nonlinear representation learning. Furthermore, although multiple transformations like that in the proposed EUDNN can be implemented by DEFE to learn deep representations, deep linear transformations are equivalent to a one layer linear representation.

In summary, DEFE cannot be employed for learning meaningful representations due to its linear nature, while the success of deep NNs is mainly caused by the meaningful representations \HL{learned} by deep nonlinear transformations, which have been explicitly implemented by the proposed EUDNN.

\section{Experiments}
\label{section_4}
In order to examine the performance of the proposed EUDNN, experiments based on a set of image classification benchmarks against selected peer competitors are performed. During the comparisons, the chosen performance metric considers the CCR on the test data. In the following, the employed benchmarks are outlined first. Then the chosen peer competitors are reviewed, and the justification \HL{for} selecting them is explained further. This is followed by the descriptions of the performance metric chosen and the specifics of parameter settings employed by these compared algorithms. Finally, the quantitative as well as the qualitative experimental results are illustrated and comprehensively analyzed.

\subsection{Benchmark Test Datasets}
Benchmarks used by compared algorithms are the handwritten digits benchmark test dataset MNIST~\cite{lecun1998gradient}, basic MNIST dataset (MNIST-basic)~\cite{larochelle2007empirical}, a rotated version of MNIST (MNIST-rot)~\cite{larochelle2007empirical}, MNIST with random noise background (MNIST-back-rand)~\cite{larochelle2007empirical}, MNIST with random image background (MNIST-back-image)~\cite{larochelle2007empirical}, MNIST-rot with random image background (MNIST-rot-back-image)~\cite{larochelle2007empirical}, tall and wide rectangles dataset (Rectangles)~\cite{larochelle2007empirical}, rectangles dataset with random image background (Rectangles-image)~\cite{larochelle2007empirical}, convex sets recognition dataset (Convex)~\cite{larochelle2007empirical}, and the gray version of Canadian Institute for Advanced Research object recognition dataset~\cite{krizhevsky2009learning}  (Cifar10-bw) over $10$ classes, i.e., airplane, automobile, bird, cat, deer, dog, frog, horse, ship, and truck.

\begin{figure}[htp]
  \centering
  \includegraphics[width=0.8\columnwidth]{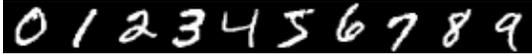}\\
  \caption{A group of digit samples ($0-9$) from the MNIST benchmark test dataset.}\label{fig_mnist_example}
\end{figure}

Briefly, these benchmark test datasets are categorized into three different classes based on the object types that they \HL{intend} to recognize. The first one is about the hand-written digits and covers the MNIST, MNIST-basic, MNIST-rot, MNIST-back-rand, MNIST-back-image, and MNIST-rot-back-image benchmarks. Examples from the MNIST benchmark are depicted in Fig.~\ref{fig_mnist_example} for reference. The second one is to classify the geometries and the rectangles, such as the Rectangles, Rectangles-image, and the Convex benchmarks. The last one is to identify the natural objects in Cifar10-bw. Different variants in MNIST and rectangles datasets present the algorithms dissimilar difficulties from the aspects of perturbations, \HL{the} small number of training dataset, and \HL{the} large testing dataset size. \HL{Furthermore, the dimensions, number of classes, and the sizes of training set and test set of the chosen benchmark datasets are shown in Table~\ref{tab_dataset_config}.}
\HL{
\begin{table}[!htp]
	\caption{\HL{The configurations of the chosen benchmark datasets.}}
	\centering
	\label{tab_dataset_config}
	\begin{tabular}{c|c|c|c|c}
		\hline
		\multirow{2}{*}{\HL{\textbf{Benchmark}}}
		&\multirow{2}{*}{\HL{\textbf{Dimension}}}&{\HL{\textbf{\# of}}}&\multicolumn{2}{c}{\HL{\textbf{Size of}}}\\
		\cline{3-5}
		&&\HL{\textbf{class}}&\HL{\textbf{training set}}&\HL{\textbf{test set}}\\
		\hline
		\HL{MNIST}& \HL{$28\times28$}& \HL{10} & \HL{50,000}& \HL{10,000}\\
		\hline
		\HL{MNIST-basic} &\HL{$28\times28$} & \HL{10}& \HL{12,000} &\HL{50,000}\\
		\hline
		\HL{MNIST-rot} &\HL{$28\times28$} &\HL{10} & \HL{12,000} &\HL{50,000} \\
		\hline
		\HL{MNIST-back-rand} & \HL{$28\times28$}&\HL{10} & \HL{12,000} &\HL{50,000} \\
		\hline
		\HL{MNIST-back-image} &\HL{$28\times28$} &\HL{10} & \HL{12,000} &\HL{50,000} \\
		\hline
		\HL{MNIST-rot-back-image} &\HL{$28\times28$} &\HL{10} & \HL{12,000} &\HL{50,000} \\
		\hline
		\HL{Rectangles} &\HL{$28\times28$} & \HL{2}& \HL{1,200}& \HL{50,000}\\
		\hline
		\HL{Rectangles-image} &\HL{$28\times28$} & \HL{2}&\HL{12,000} &\HL{50,000} \\
		\hline
		\HL{Convex} & \HL{$28\times28$}& \HL{2}& \HL{8,000}&\HL{50,000} \\
		\hline
		\HL{Cifar10-bw} &\HL{$32\times32$} &\HL{10} &\HL{50,000} & \HL{10,000}\\
		\hline

	\end{tabular}
\end{table}
}

\subsection{Performance Metric}
Technically speaking, it is difficult to directly evaluate whether the \HL{learned} representations are meaningful or not because they are intermediate outcomes. A general practice for this is to feed these \HL{learned} representations to a particular classification task, and then to investigate the CCR by a classifier. Commonly, a higher CCR implies that the \HL{learned} representations are more meaningful. Because the benchmarks employed in these experiments are multi-class classification tasks, the softmax regression classifier~\cite{engel1988polytomous} is employed here to measure the corresponding CCR according to the convention adopted in the community.

It is assumed that a set of training data and their corresponding labels with $k$ distinct integer values are denoted as $\{x_1, \cdots, x_m\}$, and $\{y_1,\cdots,y_m\}$, respectively, where $x_i\in \mathcal{R}^n$ and $y_i\in \{1,\cdots, k\}$. To be specific, the label of the sample $x_i~(i\in\{1,\cdots,m\})$ is predicted by~(\ref{eq_softmax_probability}) with the softmax regression,
\begin{equation}
\label{eq_softmax_probability}
\arg\underset{j}{\max}~~p_j(x_i)= \frac{exp(\theta_j^Tx_i)}{\sum_{l=1}^k exp(\theta_l^Tx_i)}
\end{equation}
where $\Theta=[\theta_1,\cdots, \theta_k]^T$ are obtained by minimizing
\begin{equation*}
\label{eq_softmax_minimizing}
J(\Theta) = -\frac{1}{m}
\left [
\sum_{i=1}^m\sum_{j=1}^k f(y_i, j)log\frac{exp(\theta_j^Tx_i)}{\sum_{l=1}^k exp(\theta_l^Tx_i)}
\right]
\end{equation*}
in which $f(y_i, j)=1$ if $y_i=j$, otherwise $f(y_i, j)=0$.
\subsection{Compared Algorithms}
Because of the proposed EUDNN aiming at evolving \emph{unsupervised deep neural networks} for learning \emph{meaningful representations}, algorithms related to evolving deep NNs (NEAT~\cite{stanley2002evolving}, HyperNEAT~\cite{stanley2007compositional}), unsupervised deep NNs (DBNs~\cite{hinton2006fast}, and variants of stacked AEs~\cite{bengio2007greedy}) that have been discussed in Section~\ref{section_1} should be all employed as peer competitors. However, the NEAT and the HyperNEAT cannot be used to learn meaningful representations due to the reasons that have been discussed in Section~\ref{section_1} and further analyzed in Section~\ref{section_2}. As a result, they are excluded from the selected compared algorithms. To this end, DBNs and variants of stacked AEs are employed for performing the comparison experiments. Because RBMs~\cite{smolensky1986information} and AEs~\cite{bourlard1988auto,hinton1994autoencoders,rumelhart1988learning} are the building blocks to train DBNs and stacked AEs, respectively, these two types of algorithms are considered as the peer competitors in our experiments to compare the performance of the \HL{learned} representations against that of the proposed algorithm (i.e., we will evolve RBMs and AEs as the unsupervised deep NN models, which are named EUDNN/RBM and EUDNN/AE, respectively, to perform the comparisons against considered peer competitors). Specifically, the variants of AEs, i.e., the Sparse AEs (SAEs)~\cite{olshausen1997sparse}, the Denoising EAs (DAEs)~\cite{vincent2008extracting}, and the Contractive AEs (CAEs)~\cite{rifai2011contractive}, have been proposed with different regularization terms for learning meaningful representations in recent years and also have obtained comparable performance in multiple tasks. As a consequence, they are also included as the peer competitors in the experiments, in addition to the DBNs.

\subsection{Parameter Settings}
For a fair comparison, multiple parameters in the second stage of the proposed EUDNN and the competing ones are the same. As a consequence, we will first give details of these generic parameter settings in this subsection. Then, the particular parameter settings are individually introduced. Because the best performance of the compared algorithms often strongly depends on the particular benchmark dataset and the corresponding parameter settings, in order to do a fair comparison, we first test these parameters from the range widely used in the community upon the corresponding training data, then the best performance upon test data of each compared algorithm is selected for comparisons.
\subsubsection{Learning Rate and Batch Size} The Stochastic Gradient Descent (SGD) algorithm is chosen as the algorithm to train the SAE, the DAE, the CAE, and the softmax regression, and its learning rates as well as the batch sizes vary in $\{0.0001, 0.001, 0.01, 0.1\}$ and $\{10, 100, 200\}$, respectively, according to the community convention.
\subsubsection{Number of Runs and Stop Criteria} All the compared algorithms are independently performed 30 runs. In addition, a performance monitor is injected into each epoch in training the softmax regression to record the best CCR over the test dataset as the best performance of the algorithm that feeds the 、HL{learned} representations to the softmax regression.
\subsubsection{Unit Number and Depth} The number of the units for the SAE, the DAE, the CAE, and the RBM in each layer is set to be from $200$ to $3,000$ using a $log$ function with an interval $0.5$ as recommended by~\cite{hinton2010practical}, and the maximum depth is set to be $5$ \HL{(this depth is excluded from the input layer, i.e., the maximum number of hidden layers)}.
\subsubsection{Statistical Significance} The results measured by the selected performance metric need to be statistically compared due to the heuristic natures of the first stage in the proposed EUDNN. In these experiments, the Mann-Whitney-Wilcoxon rank-sum test~\cite{steel1997principles} with a $5\%$ significant level is employed for this purpose according to the community convention.

In addition, the sparsity of the SAE, the binary corrupted level of the DAE, and the coefficient of the contractive term in the CAE are set to be $10\%, 30\%, 50\%$ and $70\%$, respectively. Because of the nature of the RBM, the CD-$k$ algorithm~\cite{carreira2005contrastive} is selected as its training algorithm and $k$ is set to be $1$ based on the suggestion in~\cite{hinton2010practical}. In order to speed up the proposed algorithm in the first stage, a proportion (i.e., $20\%$) of the training dataset is randomly selected in each generation for the fitness evaluation. In addition, the connection weights and the biases are respectively set to be between $[-4\times 6/\sqrt{n_{number}}, 4\times 6/\sqrt{n_{number}}]$ with a uniform sampling and $0$, respectively~\cite{glorot2010understanding}, if required, where $n_{number}$ denotes the total number of the units in two adjacent layers based on the experiences suggested in~\cite{glorot2010understanding}.

Because parameter settings in the second stage of the proposed EUDNN are the same as that of the peer competitors, parameter settings of the evolution related parameters in the first stage are declared next. Conveniently, one chromosome in this stage can be divided into three parts: main basis related coefficients (Part 1) which are used to represent the vector $a_1$ in Fig.~\ref{fig_encode_transformation_matrix}, projected space related coefficients (Part 2) which are employed to indicate which bases are selected for the connection weight, and the coefficients (Part 3) which denote the type of activation functions. Because Parts 1 and 2 have strong effects on the quality of the connection weight, it is hopefully that crossover operation should be promoted in these two parts for improving the exploitation local search that provides much better performance based on the exploration global search. As a consequence, one point crossover operator is employed in Parts 1 and 2. In addition, three widely used nonlinear activation functions are considered in the proposed algorithm and one is to be selected for the corresponding connection weight. Therefore, it is hopefully that the information representing the activation function is not modified often since it is hard to determine which one is the best. Consequently, Parts 2 and 3 are considered as one part to participate in the crossover operation. It is noted here that, when the value in Part 3 is invalid, a random one is chosen to reset it. \HL{Noting that the polynomial mutation~\cite{deb2001multi} is used here as the mutation operator (distribution index is set to be $20$). In addition, the population size is set to be $50$. As for the crossover probability and the mutation probability in the proposed algorithm, both of them are set to be the same as that of the community convention (i.e., $0.9$ for crossover and $0.1$ for mutation). A proportion of 10\% is randomly selected from the training set for the fitness evaluation.} Codes of the proposed EUDNN can be made available upon request through the first author.

\subsection{Experimental Results}
\label{section_quantitative_experiments}
Based on the motivation of our design, the proposed EUDNN 1) employs evolutionary algorithm and local search strategy to ensure the \HL{learned} representations through deep NNs to be meaningful, 2) employs evolutionary approach in the first stage to help the deep NNs find the optimal architectures and the good initialized weights, which give a better starting position for the second stage, and 3) employs the local search strategy in the second stage to improve the intended performance much further. Consequently, a series of experiments are carefully crafted to evaluate the performance of the proposed design.
\subsubsection{Performance of the Proposed Algorithm}
In order to quantify whether the representations \HL{learned} by the proposed EUDNN are meaningful, a series of experiments are well-designed and comparisons are performed. Specifically, EUDNN/AE and EUDNN/RBM are two implementations of the proposed algorithm over the unsupervised neural network models (i.e., AEs and RBMs, respectively). Then they are used to learn the representations together with the selected peer competitors employing the configurations introduced above. Next, the softmax regression metric is employed to measure whether the \HL{learned} representations improve the associated classification tasks through CCR, which in turn indicates the \HL{learned} representations being meaningful or not.
\begin{table*}[!htp]
  \caption{The correct classification rate of the proposed EUDNN (EUDNN/AE and EUDNN/RBM) upon MNIST, MNIST-basic, MNIST-rot, MNIST-back-rand, MNIST-back-image, MNIST-rot-back-image, Rectangles, Rectangles-image, Convex, and Cifar10-bw benchmarks against stacked denoising auto-encoder (DAE), stacked contractive auto-encoder (CAE), stacked sparse auto-encoder (SAE), and the deep belief network (DBN). Best mean values are highlighted in \HL{boldface}. The symbols ``+,'' ``-,'' and ``='' denote whether the proposed algorithm statistically are better than, worse than, and equal to that of the corresponding peer competitors, respectively, with the employed rank-sum test.}
  \centering
  \label{tab_comparison_results}
  \begin{tabular}{c|c|c|c|c|c|c}
  \hline
    \multirow{2}{*}{\textbf{Benchmark}}&\multicolumn{2}{c|}{\textbf{EUDNN}}&\multirow{2}{*}{\textbf{DAE}}&\multirow{2}{*}{\textbf{CAE}}&\multirow{2}{*}{\textbf{SAE}}&\multirow{2}{*}{\textbf{DBN}}\\
    \cline{2-3}
    &\textbf{AE}&\textbf{RBM}&&&&\\
    \hline
    MNIST&0.9878(0.00751)&\textbf{0.9885(0.00255)}&0.9820(0.00506)(+)&0.9843(0.00699)(+)&0.9832(0.00891)(+)&0.9771(0.00959)(+)\\
    \hline
    MNIST-basic&0.9674(0.00616)&0.9633(0.00473)&0.9580(0.00352)(+)&0.9635(0.00831)(+)&\textbf{0.9776(0.00585)(-)}&0.9658(0.00550)(+)\\
    \hline
    MNIST-rot&\textbf{0.7952(0.00917)}&0.7549(0.00286)&0.7274(0.00757)(+)&0.7706(0.00754)(+)&0.7852(0.00380)(+)&0.7639(0.00568)(+)\\
    \hline
    MNIST-back-rand&0.8843(0.00076)&0.8386(0.00054)&0.7725(0.00531)(+)&0.5741(0.00779)(+)&\textbf{0.8851(0.00934)(=)}&0.8221(0.00130)(+)\\
    \hline
    MNIST-back-image&0.4325(0.00569)&\textbf{0.4830(0.00469)}&0.4022(0.00012)(+)&0.4010(0.00337(+)&0.4638(0.00162)(+)&0.4587(0.00794)(+)\\
    \hline
    MNIST-rot-back-image&\textbf{0.8925(0.00906)}&0.8879(0.00815)&0.8691(0.00127)(+)&0.6574(0.00913)(+)&0.8733(0.00632)(+)&0.8830(0.00098)(=)\\
    \hline
    Rectangles&0.9627(0.00311)&\textbf{0.9681(0.00829)}&0.9232(0.00166)(+)&0.6275(0.00602)(+)&0.9408(0.00263)(+)&0.9622(0.00154)(=)\\
    \hline
    Rectangles-image&0.7521(0.00689)&0.7716(0.00048)&0.7598(0.00451)(+)&\textbf{0.7810(0.00784)(=)}&0.7725(0.00002)(-)&0.7628(0.00913)(+)\\
    \hline
    Convex&\textbf{0.8113(0.00052)}&0.8085(0.00826)&0.7930(0.00538(+)&0.8016(0.00996)(+)&0.8053(0.00878)(+)&0.7895(0.00443)(+)\\
    \hline
    Cifar10-bw&\textbf{0.4798(0.00107)}&0.4331(0.00962)&0.4309(0.00005)(+)&0.4860(0.00775)(+)&0.4423(0.00817)(+)&0.4598(0.00869)(+)\\
    \hline
    &\multicolumn{2}{c|}{+/-/=}&10/0/0&9/0/1&7/2/1&8/0/2\\
    \hline

  \end{tabular}
\end{table*}

Particularly, the mean values and standard derivations of CCR resulted by these compared algorithms over $30$ independent runs are listed in Table~\ref{tab_comparison_results} in which the best results over the same benchmark are highlighted in \HL{boldface}. In addition, the symbols ``+,'' ``-,'' and ``='' denote whether the CCR of the proposed algorithm upon the corresponding benchmarks are statistically better than, worse than, and equal to that of the associated peer competitors, respectively, with the employed rank-sum test\footnote{To do this statistically test, we first select the better CCR generated by EUDNN/AE and EUDNN/RBM with the same benchmark, then the selected results are used to do the rank-sum test.}. Furthermore, the summarizations, how many times over the considered benchmarks the proposed EUDNN are better than, worse than, and equal to the corresponding peer competitor, are listed in the last row of Table~\ref{tab_comparison_results}.

\HL{In Table~\ref{tab_best_performance_config}, the first column shows the names of the chosen benchmark datasets, the second column provides the corresponding best CCRs obtained, while the third column presents the numbers of neurons of the deep models (excluding the the classifier layer) with which the best CCRs are reached on the corresponding benchmark dataset. As we have claimed in Subsection IV-D that the maximum number of building blocks investigated in this paper is set to be five. Therefore, the number of layers, which include the input layer and hidden layers, shown in Table~\ref{tab_best_performance_config} for each benchmark dataset does not exceed six. For the first row in Table~\ref{tab_best_performance_config} as an example, it indicates that the best CCR of 98.85\% on the MNIST benchmark dataset is achieved with only four building blocks where the input layer is with 784 neurons, and hidden layers are with 400, 202, 106, and 88 neurons, respectively.}

It is clearly shown in Table~\ref{tab_comparison_results}\footnote{In this paper, the statistical results biases the results generated by the statistical significance toolkit, i.e., the Mann-Whitney-Wilcoxon rank-sum test [67] with a $5\%$ significant level.} that the proposed EUDNN/AE obtains the best mean values upon the MNIST-rot, the MNIST-rot-back-image, the Convex, and the Cifar10-bw benchmarks, and the best rank-sum results upon the MNIST-rot, the Convex, and the Cifar10-bw benchmarks. Moreover, the proposed EUDNN/RBM wins both the best mean values and the rank-sum results upon the MNIST, and the MNIST-back-image benchmarks. Although the best result of the proposed EUDNN (obtained by the EUDNN/AE) over the MNIST-basic benchmark is a little worse than that of the SAE, which is the winner of the best mean value and rank-sum results, EUDNN/AE outperforms all the other peer competitors. Furthermore, the SAE obtains the best mean values upon the MNIST-basic and the MNIST-back-rand benchmarks, but the best result of the proposed algorithm (obtained by the EUDNN/AE)  is statistically equal to that of the SAE upon the MNIST-back-rand benchmark, and also outperforms other competing algorithms. Upon the Rectangles-image benchmarks, the best result of the proposed algorithm (obtained by the EUDNN/RBM) is worse than that of the CAE and the SAE, while the EUDNN/RBM and CAE have the same results statistically. In addition, the best results of the proposed algorithm upon the MNIST-rot-back-image (obtained by the EUDNN/AE) and the Rectangles (obtained by the EUDNN/RBM) benchmarks are all statistically equivalent to that of the DBN, while the best mean values upon these two benchmarks are obtained by the EUDNN/AE and the EUDNN/RBM, respectively. Note here that the MNIST is a widely used classification benchmark for quantifying the performance of deep learning models, and the best results are frequently obtained by supervised models, which require sufficient labeled training data during their training phases. To our best knowledge, the CCR with $98.85\%$ obtained by the proposed algorithm (EUDNN/RBM), which is an unsupervised approach is a very promising result among unsupervised deep learning models. In summary, the proposed algorithm totally wins $34$ times over the $40$ comparisons against the selected peer competitors, which reveals the superior performance of the proposed algorithm in learning \emph{meaningful representations} with \emph{unsupervised neural network models}.
\subsubsection{Performance Analysis Regarding the First Stage}
Since we have claimed that the first stage of the proposed algorithm helps the unsupervised NN-based models learn optimal architectures and \HL{better-initialized} parameter values, component-wise experiments over the optimal architectures and the initialized parameter values should be performed to investigate their respective effects to justify our designs. However\HL{,} the initialized parameter values are dependent on the architectures. This leads to the specific experiment by varying only the architecture configurations on investigating how the \HL{learned} architectures solely affect the performance is difficult to design. Hence, the performance regarding the initialized parameter values is mainly investigated here.

\begin{figure}[!htp]
  \centering
  \includegraphics[width=0.8\columnwidth]{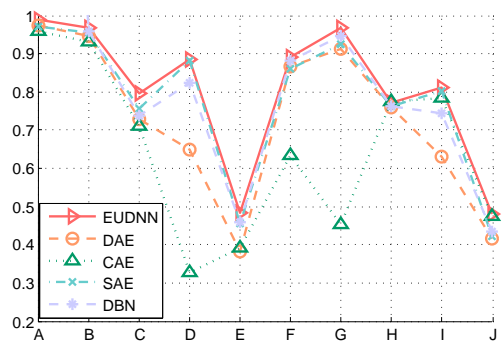}\\
  \caption{The performance of the proposed algorithm against DAE, CAE, SAE, and DBN with the configurations on which the proposed algorithm obtains the best correct classification rates over benchmarks measured by softmax regression. Especially, A-J denote the benchmarks MNIST, MNIST-basic, MNIST-rot, MNIST-back-rand, MNIST-back-image, MNIST-rot-back-image, Rectangles, Rectangles-image, Convex, and Cifar10-bw, respectively.}\label{fig_same_architecture}
\end{figure}

To this end, we first record the architecture configurations (see Table~\ref{tab_best_performance_config}) with which the proposed algorithm presents the promising performance in best mean values of EUDNN/AE and EUDNN/RBM upon each benchmark from Table~\ref{tab_comparison_results}. Then experiments are re-performed by peer competitors with the recorded architecture configurations and randomly initialized parameter values. Finally, the \HL{learned} representations are fed to the considered performance metric to measure whether these representations are meaningful. Specifically, experimental results are depicted in Fig.~\ref{fig_same_architecture} in which the vertical axis denote the CCR while A-J in the horizontal axis represent the benchmarks MNIST, MNIST-basic, MNIST-rot, MNIST-back-rand, MNIST-back-image, MNIST-rot-back-image, Rectangles, Rectangles-image, Convex, and Cifar10-bw, respectively.

It is shown in Fig.~\ref{fig_same_architecture} that most of the peer competitors employing the chosen architecture configurations listed in Table~\ref{tab_best_performance_config} obtain worse CCR upon the considered benchmarks compared to the proposed algorithm. Specifically, the proposed algorithm shows these best CCR upon MNIST, MNIST-rot, MNIST-back-image, MNIST-rot-back-image, Convex, and Cifar10-bw benchmarks, which is consistent with the findings listed in Table~\ref{tab_comparison_results}. In addition, the proposed algorithm wins the best CCR upon MNIST-basic and MNIST-back-rand benchmarks as well, with these architecture configurations. In addition to the proposed algorithm in which the initialized parameter values are set by the proposed evolutionary approach, all the results illustrated in Fig.~\ref{fig_same_architecture} are obtained by the compared algorithms with the same architecture configurations and commonly used parameter initializing methods for the second stage. As we all know that the performance of local search strategies is strongly rely on their starting position, therefore, it is reasonable to conclude that the evolutionary scheme employed by the first stage of the proposed algorithm has substantially helped the \HL{learned} representations to be meaningful.
\begin{table}[!htp]
  \caption{The best correct classification rate (CCR) of the proposed algorithm upon MNIST, MNIST-basic, MNIST-rot, MNIST-back-rand, MNIST-back-image, MNIST-rot-back-image, Rectangles, Rectangles-image, Convex, Cifar10-bw benchmarks and the corresponding architecture configurations.}
  \centering
  \label{tab_best_performance_config}
  \begin{tabular}{c|c|c}
  \hline
  \textbf{Benchmark}&\textbf{Best CCR}&\textbf{Architecture configurations}\\
  \hline
  MNIST&0.9885&784, 400, 202, 106, 88\\
  \hline
  MNIST-basic&0.9674&784, 400, 211, 120\\
  \hline
  MNIST-rot&0.7952&784, 400, 233, 133, 100, 81\\
  \hline
  MNIST-back-rand&0.8843&784, 397, 202, 123\\
  \hline
  MNIST-back-image&0.4830& 784, 386, 191, 1088, 100\\
  \hline
  MNIST-rot-back-image& 0.8925&784, 378, 205, 106\\
  \hline
  Rectangles&0.9681&784, 397, 205, 113, 100, 75\\
  \hline
  Rectangles-image&0.7716& 784, 402, 214, 122, 89\\
  \hline
  Convex&0.8113& 784, 394, 200, 110, 55, 49\\
  \hline
  Cifar10-bw&0.4798& 1024, 502, 253, 141, 130\\
  \hline
  \end{tabular}
\end{table}

\subsubsection{Performance Analysis Regarding the Second Stage}
\begin{figure}
  \centering
  \includegraphics[width=0.8\columnwidth]{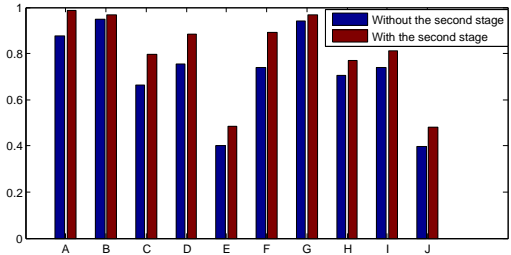}\\
  \caption{Correct classification rate (CCR) comparisons of the proposed algorithm without (denoted by blue bars) and with (denoted by red bars) the second stage upon the MNIST, MNIST-basic, MNIST-rot, MNIST-back-rand, MNIST-back-image, MNIST-rot-back-image, Rectangles, Rectangles-image, Convex, and Cifar10-bw benchmarks, which are denoted by A-J, respectively.}\label{fig_without_second_stage}
\end{figure}

In this experiment, we mainly investigate whether the local search strategy employed in the second stage promotes the integral performance of the proposed algorithm compared to only the evolutionary methods used in the first stage. For this purpose, we first pick up the promising CCR obtained by the proposed algorithm from Table~\ref{tab_comparison_results} in which the results of the proposed algorithm are collectively achieved by the evolutionary method employed in the first stage and the local search strategy employed in the second stage. Then we select the corresponding results performed without the local search strategy (i.e., the results obtained by the proposed algorithm during the first stage). Finally, these results are illustrated in Fig.~\ref{fig_without_second_stage} for quantitative comparisons. Specifically in Fig.~\ref{fig_without_second_stage} the vertical axis denotes the CCR, while A-J in the horizontal axis represent the benchmarks MNIST, MNIST-basic, MNIST-rot, MNIST-back-rand, MNIST-back-image, MNIST-rot-back-image, Rectangles, Rectangles-image, Convex, and Cifar10-bw, respectively, and the bars in blue denote the results obtained by the proposed algorithm without the second stage, while the bars in red refer to that with the second stage.

It is clearly shown in Fig.~\ref{fig_without_second_stage} that the performance has been improved with the second stage of the proposed EUDNN over all the considered benchmarks compared to the algorithm that only the first stage is employed. Especially, the CCR have been significantly improved by about $20\%$ upon the MNIST-rot, MNIST-back-rand, MNIST-back-image, MNIST-rot-back-image, and Cifar10-bw benchmarks and $12.83\%$ \HL{on} the MNIST benchmark. In summary, it is concluded from these experimental results that the local search strategy utilized in the second stage helps the performance of the proposed algorithm to be improved much further, which promotes the \HL{learned} representations to be meaningful and satisfies our motivation of this design.

\subsection{Visualizations of \HL{Learned} Representations}
\begin{figure*}[htp]
\begin{center}
\subfloat[]{\includegraphics[width=0.43\columnwidth]{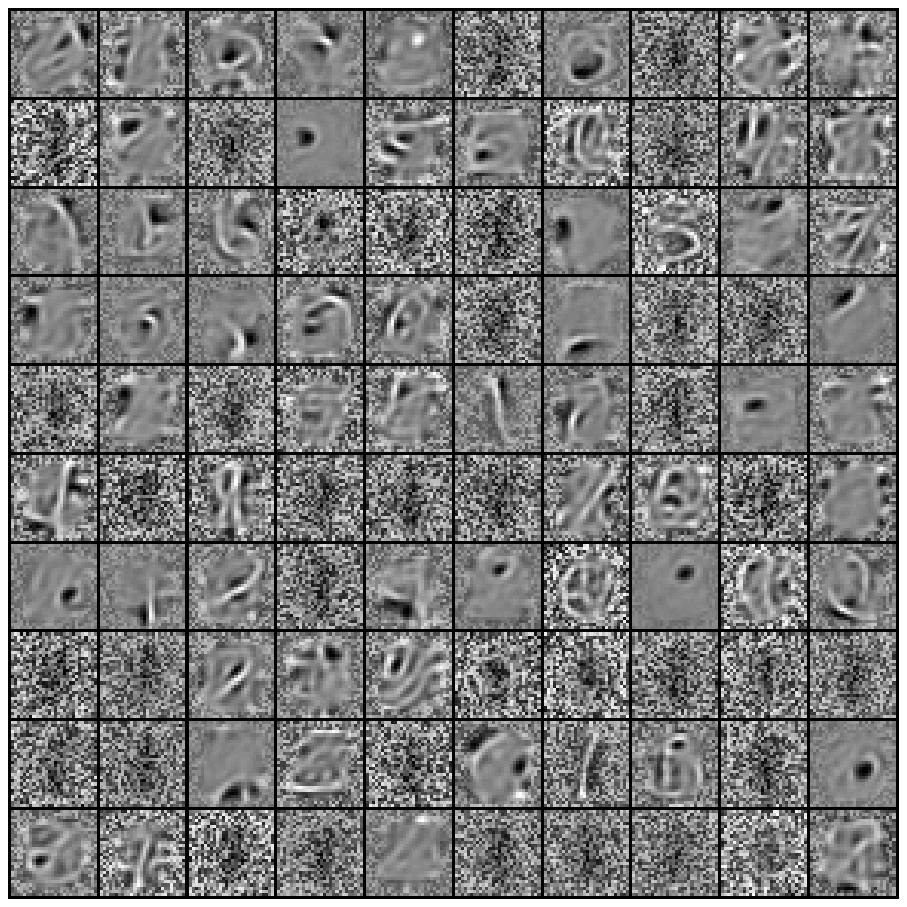}%
\label{fig_visualization_layer1}}
\hfil
\subfloat[]{\includegraphics[width=0.43\columnwidth]{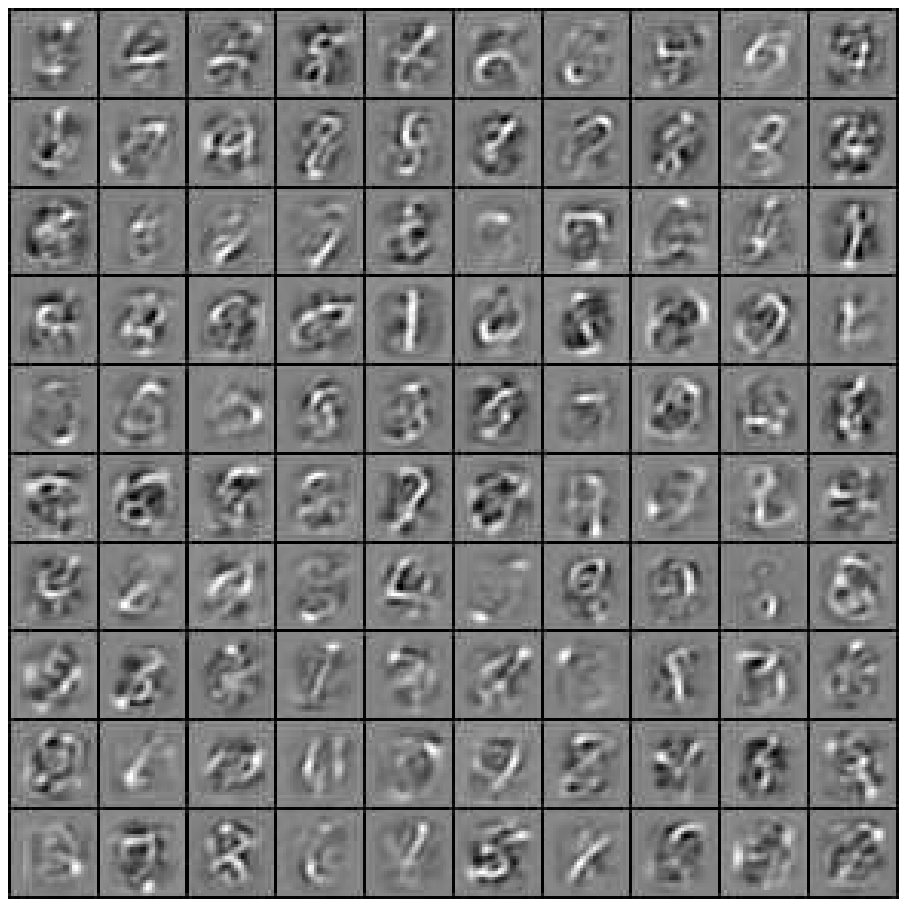}%
\label{fig_visualization_layer2}}
\hfil
\subfloat[]{\includegraphics[width=0.43\columnwidth]{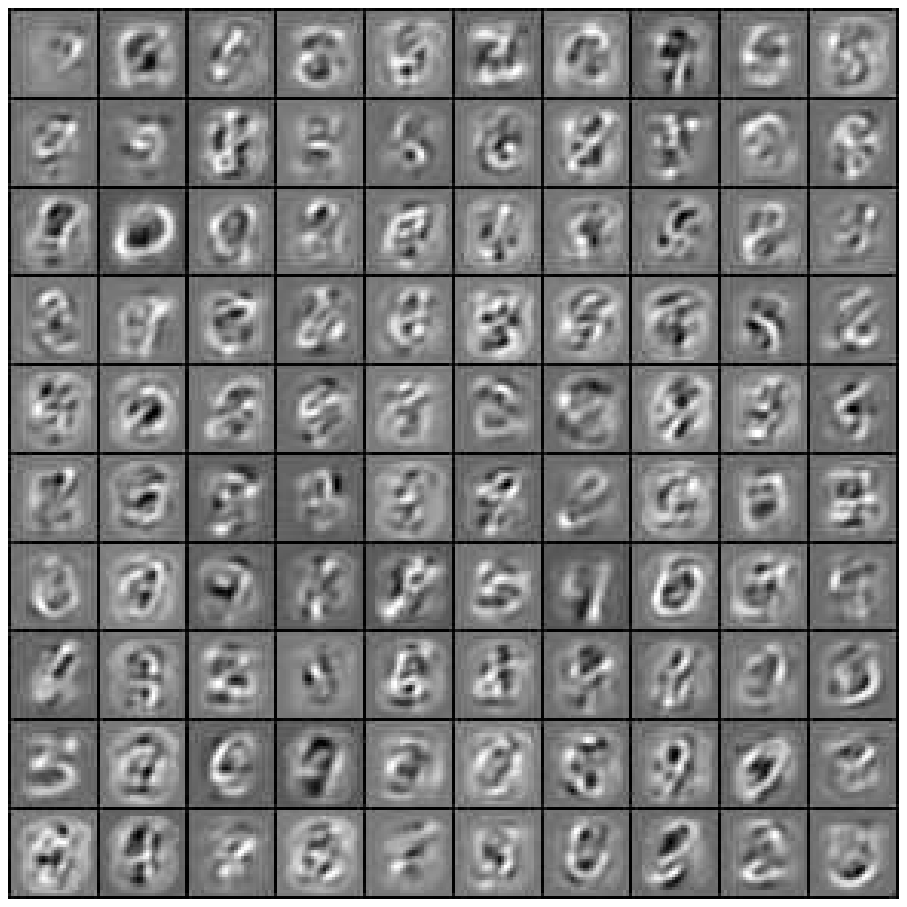}%
\label{fig_visualization_layer3}}
\caption{Visualizations of the proposed algorithm over MNIST dataset with depths $1$ (Fig.~\ref{fig_visualization_layer1}), $2$ (Fig.~\ref{fig_visualization_layer2}), and $3$ (Fig.~\ref{fig_visualization_layer3}) by activation maximization method.}
\label{fig_visualization}
\end{center}
\end{figure*}
In Subsection~\ref{section_quantitative_experiments}, a series of quantitative experiments has been given to highlight the performance of the proposed algorithm in learning meaningful representations with unsupervised deep NN-based models. Here, a qualitative experiment is provided for comprehensively understanding what the representations are \HL{learned} from the proposed algorithm via visualizations, which is a common approach employed by related works~\cite{bengio2009learning,vincent2008extracting,rifai2011contractive,sunlearning,sun2015explicit} to intuitively investigate the learning representations. For this purpose, the activation maximization approach~\cite{erhan2009visualizing} is utilized to visualize the \HL{learned} representations of the proposed algorithm over MNIST dataset and a number of $100$ randomly selected visualizations of the patches are illustrated~\footnote{Because visualizations of representations \HL{learned} from the depth larger than one are difficult to interpret, and that from the depth larger than three have no reference for comparisons, only representations with depths $1$, $2$, and $3$ are visualized here.} in Fig.~\ref{fig_visualization}. Furthermore, the SGD is employed during the optimization of the activation maximization with $10,000$ iterations and a fixed learning rate of $0.1$. To be specific, Fig.~\ref{fig_visualization_layer1} shows the \HL{learned} representations on depth $1$ in which the visualization is commonly describable~\cite{erhan2009visualizing}. It is clear in Fig.~\ref{fig_visualization_layer1} that some strokes are \HL{learned} in most patches and a part of the representations is similar to that of the RBM~\cite{erhan2009visualizing}, which can be viewed as the effectiveness of the proposed algorithm, because these similar representations over MNIST dataset have been reported in multiple \HL{kinds of literature}~\cite{sunlearning,sun2015explicit}. The visualizations of the representations on depths $2$ and $3$ are depicted in Figs.~\ref{fig_visualization_layer2} and~\ref{fig_visualization_layer3}, respectively. However, these representations are difficult to understand intuitively and be interpretable due to the high-level hierarchical nature~\cite{erhan2009visualizing}. But it still can be concluded that the proposed algorithm has \HL{learned} the meaningful representations by comparing them to the experiments simulated in~\cite{erhan2009visualizing} that \HL{learned} representations herein resemble those of the DAE to some extent. \HL{Noting that multiple learned features shown in Fig.~\ref{fig_visualization_layer1} seem to be random. The reason is that not all the neurons in the corresponding hidden layer have learned the meaningful features. Specifically, the visualization of features is from the 100 neurons randomly selected from the 313,600 (this number can be calculated from Table~\ref{tab_best_performance_config}), and it is not necessary that all the 313,600 neurons have \HL{learned} the meaningful features.} In summary, these visualizations give a qualitative observation to highlight that the meaningful representations have been effectively \HL{learned} by the proposed algorithm.
\section{Conclusion}
\label{section_5}
In order to warrant the representations \HL{learned} by \emph{unsupervised} deep neural networks to be meaningful, the existing approaches for learning them need optimal combinations of hyper-parameters, appropriate parameter values, and sufficient labeled data as the training data.
These approaches generally employ the exhaustive grid search method to directly optimize hyper-parameters due to their unavailable gradient information, which give an unaffordable computational complexity that increases with an order of magnitude as the number of hyper-parameter grows. Furthermore, the gradient-based training algorithms in these existing algorithms are \HL{easy} to be trapped into the local minima, which cannot guarantee them the best performance. In addition, in the current era of Big Data, the volume of labeled data is limited and obtaining sufficient data with labels is expensive, if not impossible. To address these concerning issues, we have proposed an evolving unsupervised deep neural networks method which heuristically searches for the best hyper-parameter settings and the global minima to learn the meaningful representations without sufficient labeled data. To be specific, two stages are composed in the proposed algorithm. In the first stage, all the information regarding hyper-parameter and parameter settings are encoded into the individual chromosome and the best one is selected when they go through a series of crossover, mutation, and selection operations. Furthermore, the activation functions that provide the nonlinear ability to the learning algorithm are also incorporated into the individual chromosome to go through the evolutions \HL{of} obtaining the promising performance. In addition, the orthogonal complementary techniques are employed in the proposed algorithm to reduce the computational complexity for effectively learning the deep representations. Specifically, only a limited number of labeled data is needed in the proposed algorithm to direct the search to learn representations with meaningfulness. For further improving the performance, the second stage is introduced with a local search strategy to complement with the ability of the exploitation search for training the proposed algorithm with the architecture and the activation function optimized in the first stage. These two stages collectively promote the proposed algorithm effectively learning the meaningful representations with unsupervised deep neural network-based models. To evaluate the meaningfulness of the \HL{learned} representations, a series of experiments are given against peer competitors over multiple benchmarks related classification tasks. The results measured by the softmax regression show the considerable competitiveness of the proposed algorithm in learning meaningful representations. In near future, we will place more focus on the efficient encoding methods as well as the way measuring the quality of the representation during the evolution \HL{of} a larger scale and higher dimensional data. In addition, we would also investigate how to effectively evolve deep supervised neural networks, such as CNNs.

\ifCLASSOPTIONcaptionsoff
  \newpage
\fi

\begin{IEEEbiography}[{\includegraphics[width=1in,height=1.25in,clip,keepaspectratio]{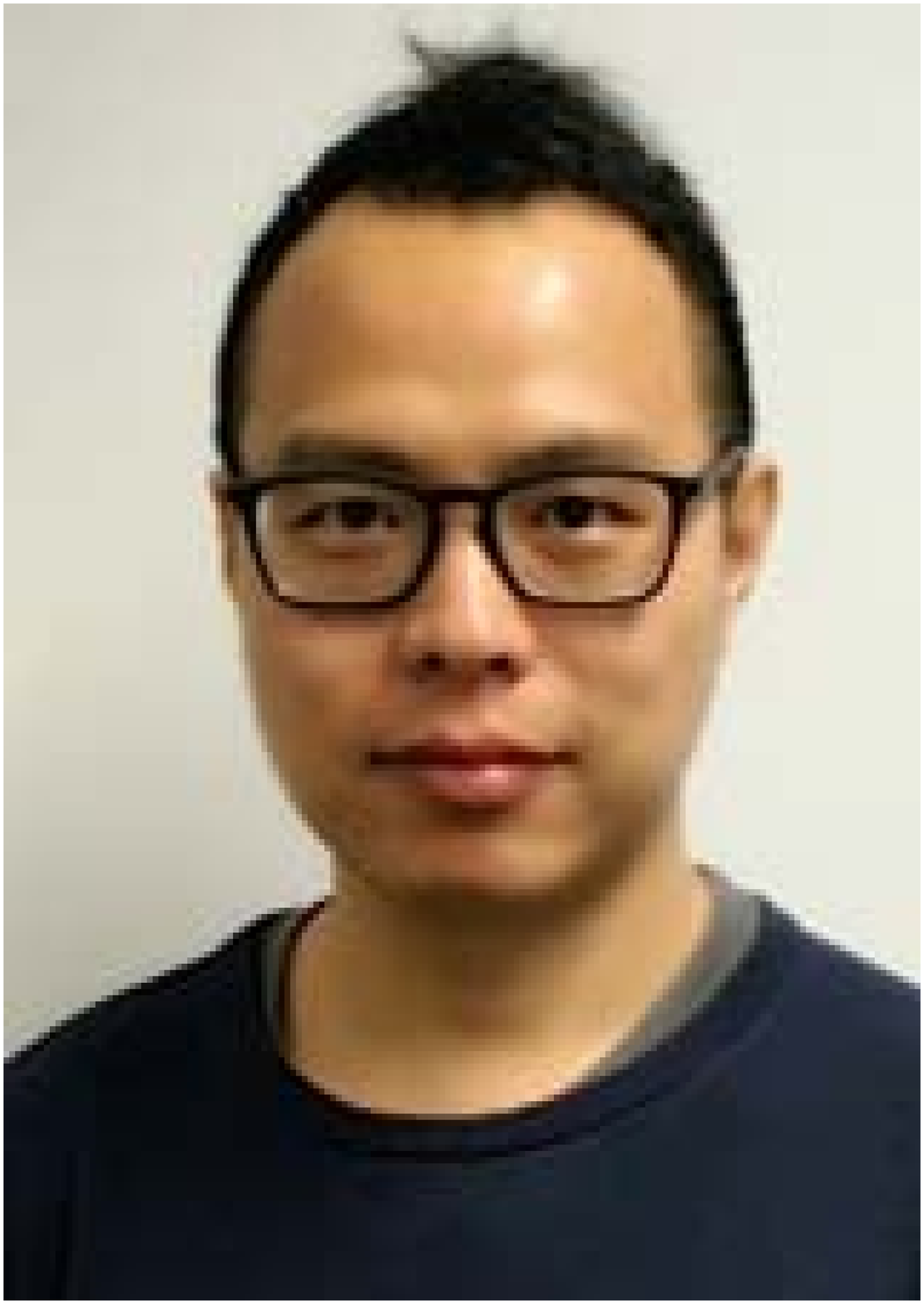}}]{Yanan Sun} (S'15-M'18) received a Ph.D. degree in engineering from the Sichuan University, Chengdu, China, in 2017. From 2015.08-2017.02, he is a jointly Ph.D. student financed by the China Scholarship Council in the School of Electrical and Computer Engineering, Oklahoma State University (OSU), USA. He is currently a Postdoc Research Fellow in the School of Engineering and Computer Science, Victoria University of Wellington, Wellington, New Zealand. His research topics are many-objective optimization and deep learning.
\end{IEEEbiography}

\begin{IEEEbiography}[{\includegraphics[width=1in,height=1.25in,clip,keepaspectratio]{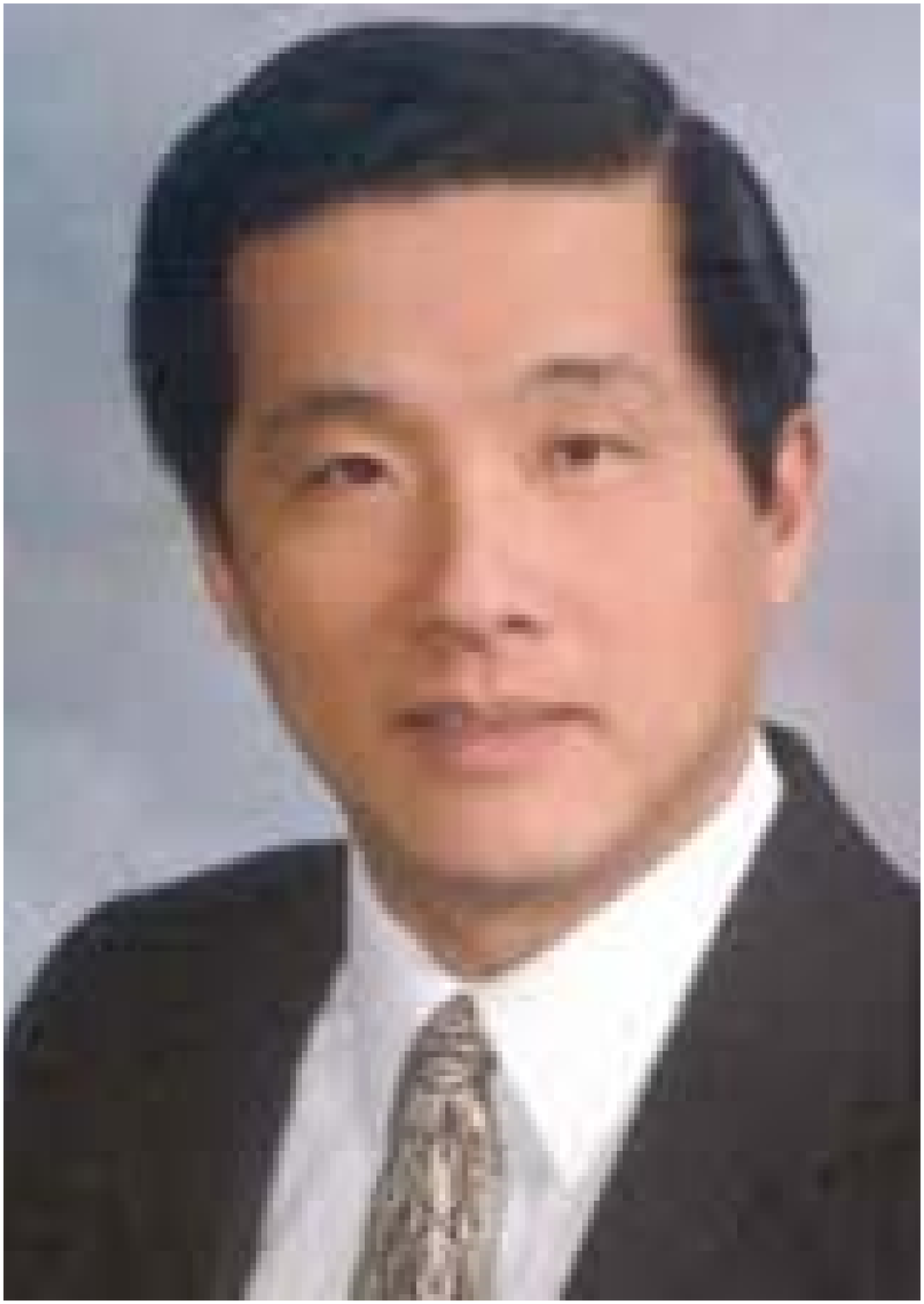}}]{Gary G. Yen}
	(S'87-M'88-SM'97-F'09) received a Ph.D. degree in electrical and computer engineering from the University of Notre Dame in 1992. Currently he is a Regents Professor in the School of Electrical and Computer Engineering, Oklahoma State University (OSU). Before joined OSU in 1997, he was with the Structure Control Division, U.S. Air Force Research Laboratory in Albuquerque. His research interest includes intelligent control, computational intelligence, conditional health monitoring, signal processing and their industrial/defense applications.
	
	Dr. Yen was an associate editor of the \textit{IEEE Control Systems Magazine, IEEE Transactions on Control Systems Technology}, \textit{Automatica}, \textit{Mechantronics}, \textit{IEEE Transactions on Systems, Man and Cybernetics, Parts A and B} and I\textit{EEE Transactions on Neural Networks}. He is currently serving as an associate editor for the \textit{IEEE Transactions on Evolutionary Computation} and the \textit{IEEE Transactions on Cybernetics}. He served as the General Chair for the \textit{2003 IEEE International Symposium on Intelligent Control} held in Houston, TX and \textit{2006 IEEE World Congress on Computational Intelligence} held in Vancouver, Canada. Dr. Yen served as Vice President for the Technical Activities in 2005-2006 and then President in 2010-2011 of the IEEE Computational intelligence Society. He was the founding editor-in-chief of the \textit{IEEE Computational Intelligence Magazine}, 2006-2009. In 2011, he received Andrew P Sage Best Transactions Paper award from \textit{IEEE Systems, Man and Cybernetics Society} and in 2014, he received Meritorious Service award from \textit{IEEE Computational Intelligence Society}.
\end{IEEEbiography}

\begin{IEEEbiography}[{\includegraphics[width=1in,height=1.25in,clip,keepaspectratio]{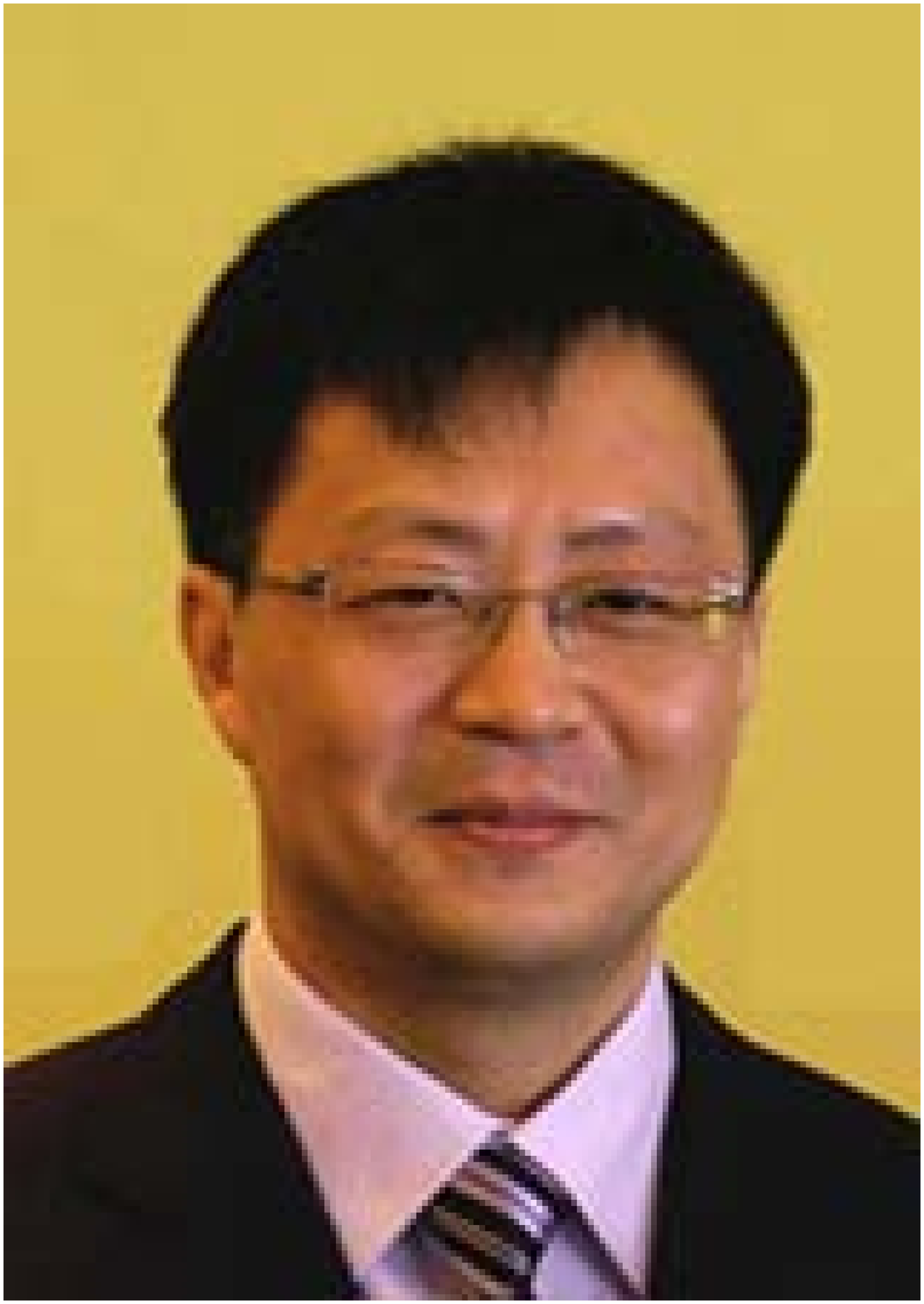}}]{Zhang Yi} (F'16)
	received a Ph.D. degree in mathematics from the Institute of Mathematics, The Chinese Academy of Science, Beijing, China, in 1994. Currently, he is a Professor at the Machine Intelligence Laboratory, College of Computer Science, Sichuan University, Chengdu, China. He is the co-author of three books: \emph{Convergence Analysis of Recurrent Neural Networks} (Kluwer Academic Publishers, 2004), \emph{Neural Networks: Computational Models and Applications} (Springer, 2007), and \emph{Subspace Learning of Neural Networks} (CRC Press, 2010). He was an Associate Editor of \emph{IEEE Transactions on Neural Networks and Learning Systems} (2009~2012), and He is an Associate Editor of \emph{IEEE Transactions on Cybernetics} (2014~). His current research interests include Neural Networks and Big Data. He is a fellow of IEEE.
\end{IEEEbiography}

\section*{Supplemental Materials}
\label{appendix_1}
\section{An Example of Encoding Strategy}
In this section, the main steps of the encoding scheme are given, and then an illustrating example based on these steps is provided. For the convenience of the development, assuming the input data is with the dimension of $n\times n$. The main steps of this encoding scheme are detailed below:
\begin{enumerate}
	\item Randomly generate $n$ orthogonal vectors and each vector is $n$-dimensional, these vectors are denoted by $S=[s_1,\cdots, s_n]$;
	\item Randomly generate $n$ real numbers that are denoted by $b=[b_1,\cdots, b_n]$;
	\item Compute $a_1=b_1\times s_1 + \cdots + b_n\times s_n$;
	\item 	Compute the bases $a_2,\cdots, a_n$ of the null space of $a_1$;
	\item 	Initialize a chromosome with a length of $2n+1$;
	\item 	Copy the values of $b_1,\cdots, b_n$ into the first position to the $n$-th position of this chromosome (i.e., they are used to denote the value of $a_1$);
	\item 	Randomly generate $n-1$ numbers from $\{0,1\}$, copy them to the $(n+1)$-th to $(2n-1)$-th position of this chromosome (they are used to represent whether the corresponding basis from $\{a_2,\cdots,a_n \}$ would be selected as the subspace or not);
	\item 	Randomly generate a number from $\{1,2,3\}$, convert it to the binary format with the length of 2, and copy it to the $2n$-th to the $(2n+1)$-th position of this chromosome (they are used to denote the index of the chosen activation function).
\end{enumerate}

Supposed that $n$ is equal to $5$, an example based on the description above is given as follow.
\begin{enumerate}
	
	\item Randomly generate vectors $S=[s_1, s_2, s_3, s_4, s_5]$ where $S\in R^{5\times 5}$;
	
	\begin{equation*}  \footnotesize{
		\begin{array}{rl}
		S=& \left[
		\begin{array}{rrrrr}
		-0.4861  & -0.6498 &   0.2718  &  0.1572 &   0.4927 \\
		-0.4617  & -0.2830 &  -0.1205  & -0.6073 &  -0.5686 \\
		-0.4438  &  0.3468 &   0.5339  &  0.4669 &  -0.4240 \\
		-0.4721  &  0.6142 &  -0.0597  & -0.3831 &   0.4995 \\
		-0.3614  &  0.0075 &  -0.7893  &  0.4916 &  -0.0681 \\
		\end{array}
		\right]\\
		\end{array}}
	\end{equation*}
	
	\item Randomly generate $5$ numbers stored into $b$;
	\begin{equation*}  \footnotesize{
		\begin{array}{rl}
		b=& \left[
		\begin{array}{ccccc}
		0.7303  &  0.4886  &  0.5785  &  0.2373  &  0.4588 \\
		\end{array}
		\right]\\
		\end{array}}
	\end{equation*}
	
	\item Compute the linear combination $a_1$ of $S$ and $b$;
	\begin{equation*}\footnotesize{
		\begin{array}{rl}
		a_1=& \left[
		\begin{array}{ccccc}
		1.2642 &  -1.4589 &  -1.0880  &  1.2815 &  -0.1746 \\
		\end{array}
		\right]^T\\
		\end{array}}
	\end{equation*}
	\item
	Compute the bases ($a_2, a_3, a_4,$ and $a_5$) of the null space of $a_1$;
	\begin{equation*}\footnotesize{
		\begin{array}{rl}
		a_2=& \left[
		\begin{array}{ccccc}
		-0.8108  &  0.4590 &   0.0400  &  0.0336 &  -0.3596 \\
		\end{array}
		\right]^T\\
		\end{array}}
	\end{equation*}
	\begin{equation*}\footnotesize{
		\begin{array}{rl}
		a_3=& \left[
		\begin{array}{ccccc}
		0.0600  &  0.0400  &  0.9970  & -0.0025  &  0.0266 \\
		\end{array}
		\right]^T\\
		\end{array}}
	\end{equation*}
	\begin{equation*}\footnotesize{
		\begin{array}{rl}
		a_4=& \left[
		\begin{array}{ccccc}
		0.0504  &  0.0336 &  -0.0025  &  0.9979  &  0.0224 \\
		\end{array}
		\right]^T\\
		\end{array}}
	\end{equation*}
	\begin{equation*}\footnotesize{
		\begin{array}{rl}
		a_5=& \left[
		\begin{array}{ccccc}
		-0.5388 &  -0.3596 &   0.0266  &  0.0224  &  0.7611 \\
		\end{array}
		\right]^T\\
		\end{array}}
	\end{equation*}

	\item Initialize a chromosome with a length of $11$;
	\begin{figure}[htp]
		\centering
		\includegraphics[width=0.8\columnwidth]{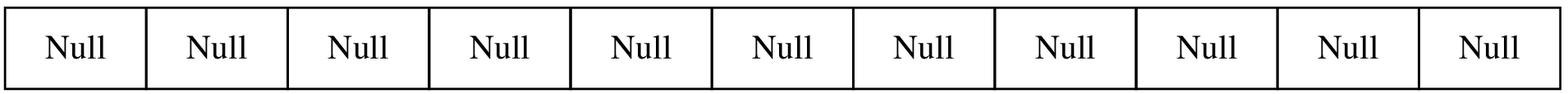}\\
	\end{figure}
	
	\item
	Copy the elements in $b$ into the $1^{\text{rd}}$---$5^{\text{th}}$ positions.
	\begin{figure}[htp]
		\centering
		\includegraphics[width=0.8\columnwidth]{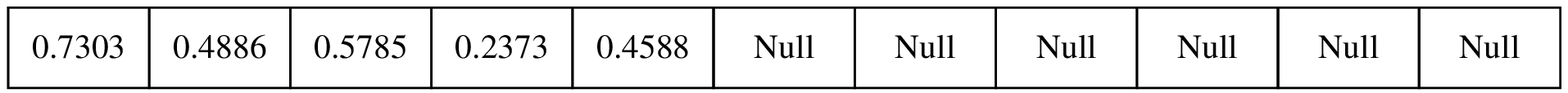}\\
	\end{figure}
	\item
	Copy $\{0, 1, 1, 0\}$ that are 4 randomly generated numbers from $\{0, 1\}$ into the $6^{\text{th}}$---$9^{\text{th}}$ positions.
	\begin{figure}[htp]
		\centering
		\includegraphics[width=0.8\columnwidth]{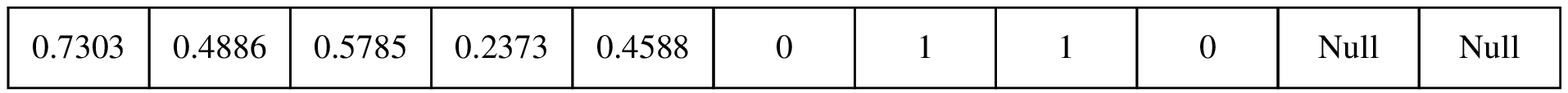}\\
	\end{figure}
	
	\item
	A randomly generated number 2 from $\{1, 2, 3\}$, convert $2$ to its binary form $10$, and copy $10$ to the $10^{\text{th}}$---$11^{\text{th}}$ positions.
	\begin{figure}[htp]
		\centering
		\includegraphics[width=0.8\columnwidth]{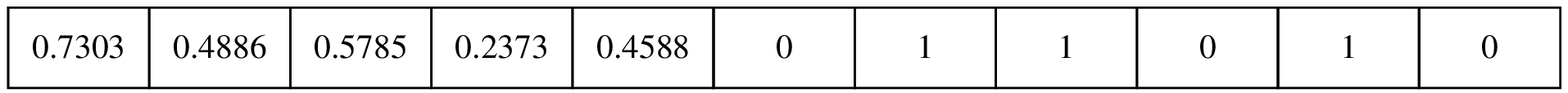}\\
	\end{figure}
\end{enumerate}

\section{Steps of Crossover and Mutation}
In the following, we will give the steps of the crossover operation on two parent solutions in the proposed algorithm.
\begin{enumerate}
	\item Assuming that the parent solutions are denoted by $ind_1$ and $ind_2$;
	\item Randomly generate a number from $[0,1]$, and if the generated number is below the predefined crossover probability, perform steps 3)-5), otherwise go to step 6);
	\item	Calculate the length (denote by $l_1$) of the first two parts, and the length (denoted by $l_2$) of the third part of the individual (the information of these three parts can be seen in Section IV-D of the manuscript);
	\item  Randomly generate an integer number (denoted by $i_1$) from $[1,l_1]$, and another integer number (denoted by $i_2$) from $[1,l_2]$;
	\item Exchange the first two parts of $ind_1$ and $ind_2$ on the position $i_1$ with the one point crossover operator, and exchange the third part of $ind_1$ and $ind_2$ on the position $i_2$ with the one point crossover operator;
	\item Return $ind_1$ and $ind_2$.
\end{enumerate}

Next, we will give the steps of the mutation operation on the individual $ind_1$.
\begin{enumerate}
	\item Randomly generate a number from $[0,1]$, and if this number is below the predefined mutation probability, performed steps 2), otherwise go to step 3);
	\item For each position in $ind_1$, randomly generate a number from $[0,1]$, if this number is less than $0.5$, perform the polynomial mutation on the current position, otherwise skip to next position.
	\item Return $ind_1$.
\end{enumerate}

\end{document}